\documentclass[10pt,twocolumn,letterpaper]{article}

\usepackage[pagenumbers]{cvpr} 
\usepackage{color}

\usepackage{pifont}
\newcommand{\cmark}{\ding{51}}%
\newcommand{\xmark}{\ding{55}}%
\usepackage{multirow}
\usepackage{multicol}
\usepackage{booktabs}
\usepackage{wrapfig}
\usepackage{tabularx}

\usepackage[dvipsnames,table]{xcolor}

\newcolumntype{Y}{>{\centering\arraybackslash}X}

\definecolor{cvprblue}{rgb}{0.21,0.49,0.74}
\usepackage[pagebackref,breaklinks,colorlinks,citecolor=cvprblue]{hyperref}

\def\paperID{6580} 
\def\confName{CVPR}
\def\confYear{2024}

\title{ZeroRF: Fast Sparse View 360$^{\circ}$ Reconstruction with Zero Pretraining}

\author{Ruoxi Shi\textsuperscript{*} \quad Xinyue Wei\textsuperscript{*} \quad Cheng Wang \quad Hao Su \\
UC San Diego
}

\begin{document}


\def\paper{1}
\def\supp{1}

\if\paper1

\twocolumn[{
\maketitle
    \vspace{-2.5em}
\begin{center}
    \includegraphics[width=0.97\linewidth]{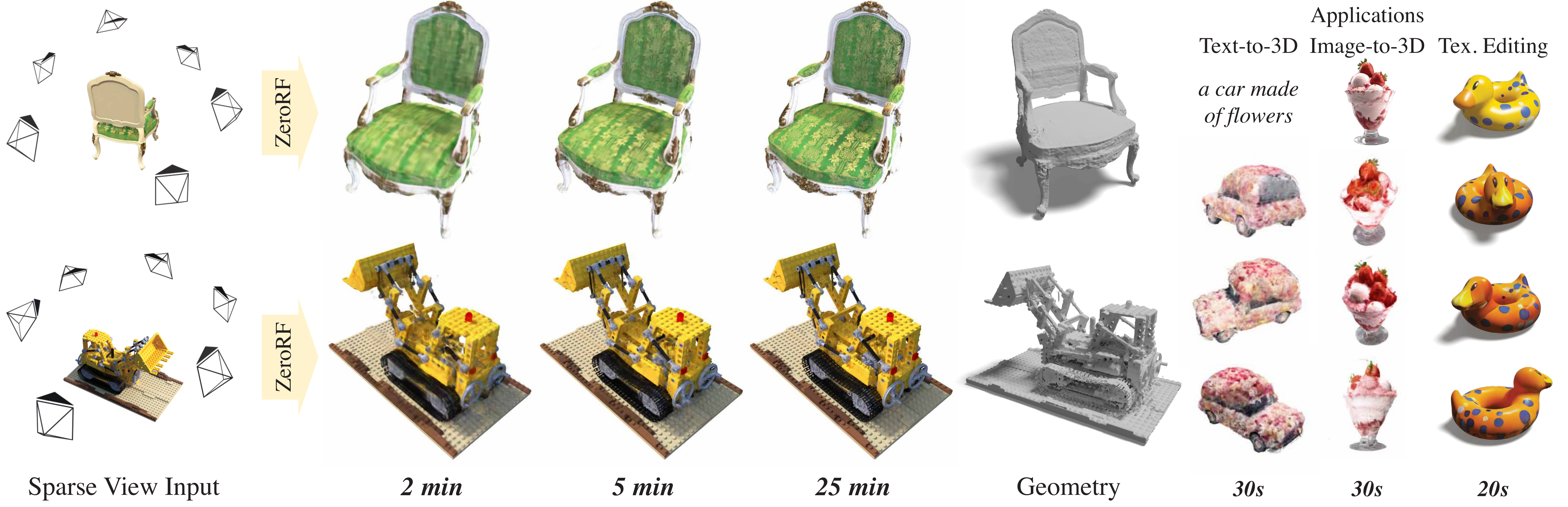}
    \label{fig:teaser}
    \vspace{-0.2em}
    \captionof{figure}{We demonstrate fast 360$^\circ$ reconstruction from sparse training views via ZeroRF. ZeroRF is able to perform novel view synthesis from few views (6 as shown in the figure) with exceptional quality, while also being fast, obtaining competitive results within 2 minutes and finishing in around 25 minutes at the full $800^2$ resolution. For common resolutions like $256^2$ or $320^2$ in 3D generation applications, ZeroRF reconstructs an object from sparse-view generations in only 30 seconds.}
\end{center}
}]
\footnotetext[1]{* Equal contribution.}

\begin{abstract}

We present ZeroRF, a novel per-scene optimization method addressing the challenge of sparse view 360° reconstruction in neural field representations. Current breakthroughs like Neural Radiance Fields (NeRF) have demonstrated high-fidelity image synthesis but struggle with sparse input views. Existing methods, such as Generalizable NeRFs and per-scene optimization approaches, face limitations in data dependency, computational cost, and generalization across diverse scenarios.
To overcome these challenges, we propose ZeroRF, whose key idea is to integrate a tailored Deep Image Prior into a factorized NeRF representation. Unlike traditional methods, ZeroRF parametrizes feature grids with a neural network generator, enabling efficient sparse view 360° reconstruction without any pretraining or additional regularization. Extensive experiments showcase ZeroRF's versatility and superiority in terms of both quality and speed, achieving state-of-the-art results on benchmark datasets.
ZeroRF's significance extends to applications in 3D content generation and editing.
 Project page: \url{https://sarahweiii.github.io/zerorf/}.
\end{abstract}    
\section{Introduction}
Breakthroughs in neural field representations, like Neural Radiance Fields (NeRF)~\cite{mildenhall2021nerf} and its subsequent developments~\cite{fridovich2022plenoxels,sun2022direct,verbin2022ref,muller2022instant,chen2022tensorf,barron2022mip,zhang2020nerf++,chen2023factor,chen2023dictionary,chen2022mobilenerf,tang2022nerf2mesh,wei2023neumanifold}, have paved the way for high-fidelity image synthesis, expedited optimization processes, and various downstream applications. Nevertheless, these approaches hinge on having a rich set of input views, and they exhibit a marked degradation in performance when confronted with sparse input views. In practical scenarios, it is not always feasible to obtain a comprehensive set of high-resolution images along with precise camera data, especially when it comes to 3D content generation \cite{liu2023one,long2023wonder3d,shi2023zero123++}. Therefore, addressing the reconstruction from sparse views presents a notable challenge, yet it remains a critical and pivotal area of interest.

In recent years, there has been a growing focus on methods tailored for sparse view reconstruction~\cite{yu2021pixelnerf,ibrnet,chen2021mvsnerf,long2022sparseneus,niemeyer2022regnerf,jain2021putting,kim2022infonerf,wang2023sparsenerf,truong2023sparf}. One line of approaches~\cite{yu2021pixelnerf,chen2021mvsnerf,long2022sparseneus,li2023instant}, commonly referred to as \emph{Generalizable NeRFs}, rely on extensive pretraining with substantial time and data requirements to directly reconstruct the scenes of interest. Performances of these models are thus closely related to the quality of the training data, and their resolutions are limited due to the heavy computation cost of large neural networks. Moreover, it is also hard for these models to generalize effectively across diverse scenarios. Other approaches that follow the per-scene optimization paradigm incorporate extra modules, like vision language models~\cite{jain2021putting} and depth estimators~\cite{wang2023sparsenerf} to help with the reconstruction. While these methods prove effective in managing narrow baselines, they fall short in achieving optimal performance in 360° reconstruction. Additionally, their applicability to real-world data is limited due to their dependence on additional supervision, which may not always be available or accurate. People have also manually designed priors spanning continuity~\cite{niemeyer2022regnerf}, information theory~\cite{kim2022infonerf}, symmetry~\cite{seo2023flipnerf} and frequency~\cite{Yang2023FreeNeRF} regularizations for the task. However, the extra regularizations may prevent the NeRFs from reconstructing the scenes faithfully~\cite{Yang2023FreeNeRF}. Furthermore, handcrafted priors are often not robust to even quite subtle setting changes.

We also observe that existing per-scene optimization approaches for 360$^\circ$ reconstruction typically demand hours of training even on the most powerful GPUs today, which hinders their use in real applications. All of them are based on the original NeRF representation, which converges much slower compared to factorized NeRF representations like Instant-NGP \cite{muller2022instant} or TensoRF \cite{chen2022tensorf}. The reason is that those handcrafted priors can hardly be applied to new representations. FreeNeRF \cite{Yang2023FreeNeRF}, for example, uses a regularization technique upon positional encodings that are specific to NeRF.

\begin{figure}[tb]
\begin{center}
   \includegraphics[width=\linewidth]{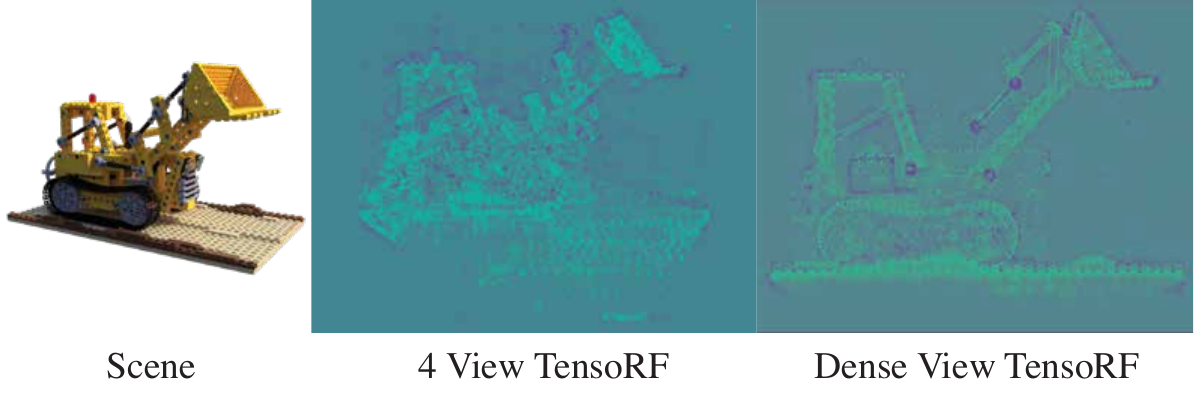}
\end{center}
\vspace{-1.5em}
  \caption{\textbf{Visualization of features obtained by fitting a vanilla TensoRF on sparse and dense views.} With dense views the features obtained are clean, while with sparse views the features are distorted with lots of noise and unwanted artifacts.}
\label{fig:motiv}
\end{figure}

We fit a TensoRF~\cite{chen2022tensorf} with different number of training views (4 and 100) on the Lego scene from the NeRF-Synthetic~\cite{mildenhall2021nerf} dataset and visualize one channel from the plane features after the training converges. From Fig.~\ref{fig:motiv} we can clearly see ray artifacts that result in noisy and distorted features under the sparse (4) view setting, while with dense (100) views the feature plane looks exactly like an orthogonal projection image of the Lego. We carried out similar experiments on the triplane~\cite{chan2021eg3d,fridovich2023k} and Dictionary Fields~\cite{chen2023dictionary} representations and find that this is not specific to TensoRF but is a general phenomenon for these grid-based factorized representations. Thus, we hypothesize that \emph{fast sparse view reconstruction with optimization can be achieved if the factorization features remain clean under sparse view supervision}.

To verify and achieve this, we propose to integrate a tailored version of the Deep Image Prior~\cite{ulyanov2018deep} into a factorized NeRF representation (See Fig.~\ref{fig:pipeline}). Instead of directly optimizing feature grids as in TensoRF, K-planes or Dictionary Fields~\cite{chen2022tensorf,fridovich2023k,chen2023dictionary}, we parametrize the feature grids with a randomly-initialized deep neural network (\emph{generator}). The intuition behind this is that with under-determined supervision, neural networks generalize much better than look-up grids for the vast majority of cases, if not always. More theoretically speaking, neural networks have much higher impedance on noise and artifacts compared to data easy to perceive and remember~\cite{ulyanov2018deep,heckel2018deep,heckel2019denoising}. The design works without any extra regularizations or pretraining, and can uniformly apply to multiple representations. The parametrization is also ``lossless'' as there exist a set of deep network parameters such that any given target feature grid could be achieved~\cite{ulyanov2018deep}.

We carried out extensive experiments on different generator networks for parametrization and different factorized representations to find the most suitable combinations for sparse view 360$^\circ$ reconstruction (Sec.~\ref{sec:analysis}), and come up with ZeroRF, a novel per-scene optimization method for this challenging task. ZeroRF 1) {does not require any sort of model pretraining}, avoiding any potential bias towards training data and any limits on settings like resolution or camera distribution; 2) {is fast in training and inference}, as it is built upon factorized NeRF representations, running in as low as 30 seconds; 3) {has the same theoretical expressiveness as the underlying factorized representations}; 4) {achieves state-of-the-art quality} for novel view synthesis with sparse-view input on NeRF-Synthetic~\cite{mildenhall2021nerf} and OpenIllumination~\cite{liu2023openillumination} benchmarks (Sec.~\ref{sec:results}).

Given the high-quality 360° reconstruction capabilities of ZeroRF, our method finds applications in various domains, including 3D content generation and editing. The potential of our approach in addressing these tasks is demonstrated in Sec.~\ref{sec:app}.

\section{Related Work}

\subsection{Novel View Synthesis}
Neural rendering techniques have paved the way for achieving photo-realistic rendering quality in novel view synthesis. It all started with the inception of Neural Radiance Field (NeRF)~\cite{mildenhall2021nerf}, which was the first to introduce a Multilayer Perceptron (MLP) for storing the radiance field and achieving remarkable rendering quality through volume rendering. Subsequently, many follow-up studies have presented various representations aimed at further enhancing performance. For instance, approaches like Plenoxels~\cite{fridovich2022plenoxels} and DVGO~\cite{sun2022direct} employed voxel-based representations, while TensoRF~\cite{chen2022tensorf}, instant-NGP~\cite{muller2022instant}, and DiF~\cite{chen2023dictionary} put forward decomposition strategies to expedite training. MipNeRF~\cite{barron2022mip} and RefNeRF~\cite{verbin2022ref} are founded on coordinate-based MLPs, and Point-NeRF~\cite{xu2022point} relies on a point-cloud-based representation. 

Some methods replace the density field with the Signed Distance Function (SDF)~\cite{wang2021neus,wang2022neus2,yariv2021volume,Oechsle2021ICCV,li2023neuralangelo,rosu2023permutosdf} or turn density fields into mesh representation ~\cite{chen2022mobilenerf, munkberg2022extracting, tang2022nerf2mesh, wei2023neumanifold, yariv2023bakedsdf} to improve surface reconstruction. These methods can extract superior-quality meshes without a substantial compromise in their rendering quality. Additionally, recent works~\cite{kerbl20233d,wu20234d,yang2023deformable} have used Gaussian splatting to achieve real-time radiance field rendering.

\subsection{Deep Network Priors}

While people commonly believe that the success of deep neural networks is due to their capability to learn from large-scale datasets, the architecture of deep networks actually capture a great amount of features prior to any learning. Training a linear classifier on features from a random convolutional network can yield performance much higher than random guess~\cite{grill2020bootstrap}. Features from randomly initialized networks are also good for few-shot learners~\cite{amid2022learning,gaier2019weight,sanghi2020powerful}. Via distillation upon this random features, the prior can be pushed further, with a line of self-supervised methods including BYOL~\cite{grill2020bootstrap}, DeepCluster~\cite{caron2018deep} and Selective Pseudo-labeling~\cite{mahon2021selective} starting from this inductive bias and use different methods to boost this prior for representation learning for images.

In contrast to these works, Deep Image Prior~\cite{ulyanov2018deep} directly exploits this deep prior without further distillation. It shows that a GAN generator architecture can act as a parametrization with high noise impedence, and thus can be applied to image restoration tasks such as denoising, super-resolution and inpainting. This is further applied to various imaging and microscopy applications ~\cite{ongie2020deep,shen2022nerp,shamshad2023transformers,lustig2007sparse,van2018compressed}, and extended with theoretical and practical improvements in Deep Decoders \cite{heckel2018deep,heckel2019denoising}. ZeroRF follows a similar paradigm to embed the deep prior into the parametrization of radiance fields.

\subsection{Sparse View Reconstruction}
Despite of the exceptional performance, NeRF models exhibit limitations in producing accurate solutions when trained with sparse observations due to insufficient information. To address this challenge, some methods opt for \textit{pretraining}~\cite{chibane2021stereo,chen2021mvsnerf,rematas2021sharf,trevithick2021grf,yu2021pixelnerf} on extensive datasets to impart prior knowledge and fine-tune the model on the target scene. Conversely, an alternative line of research focuses on per-scene optimization through \textit{manually designed regularizations}~\cite{roessle2022dense,jain2021putting,seo2023mixnerf,kwak2023geconerf,truong2023sparf,seo2023flipnerf,Yang2023FreeNeRF}. For example, to increase semantic consistency, DietNeRF~\cite{jain2021putting} extracts high-level features with the CLIP Vision Transformer~\cite{radford2021learning}. Many of them design loss functions to alleviate cross-view inconsistency, either based on information theory~\cite{kim2022infonerf, niemeyer2022regnerf}. SPARF~\cite{truong2023sparf,wang2023sparsenerf,deng2022depthsupervised} leverages pretrained networks for correspondence or depth estimation to compensate for the lack of 3D information. Different from these existing arts, ZeroRF demonstrates a remarkable ability to synthesize novel views without relying on pretraining or explicit regularizations.

\section{Preliminaries}
\textbf{Neural Radiance Field (NeRF)} represents a 3D scene radiance field by an MLP, where given an input 3D location $x$ and the view direction $d$, it outputs the volume density $\sigma_x$ and view-dependent color $c_x$:

\begin{equation}
    \sigma_{x}, c_x = F(x,d)
\end{equation}

Then the density $\sigma$ and color $c$ are used in the differentiable volume rendering:

\begin{equation}
\small
\centering
        \hat{C}(r) =  \sum^N_{i=1} T_i \left(1-\exp \left(-\sigma_i \delta_i\right)\right) c_i,
        T_i = \exp \left(-\sum_{j=1}^{i-1} \sigma_j \delta_j\right)
    \label{eq:raymarching}
\end{equation}
where $\hat{C}(r)$ is the volume rendering predicted RGB colors for ray $r$, $T$ is the volume transmittance and $\delta$ is the ray marching step size. The whole rendering process is differentiable, which allows the neural network to be optimized by rendering loss:

\begin{equation}
\small
\centering
        \mathcal{L} = \sum_{r \in R} || \hat{C}(r) - C(r)||_2^2
    \label{eq:rendering_loss}
\end{equation}
where $C(r)$ is the ground truth RGB colors.

\begin{figure*}[th]
\begin{center}
   \includegraphics[width=\linewidth]{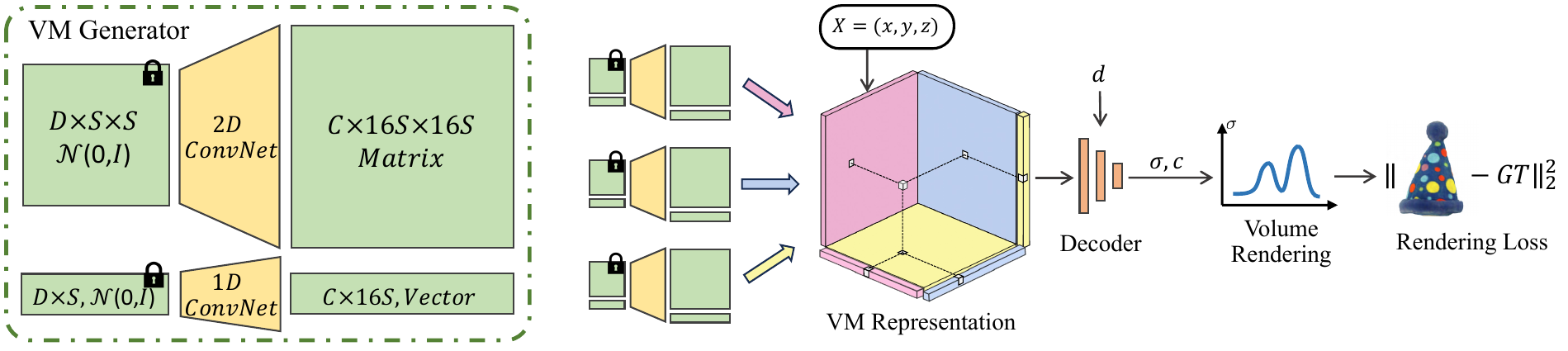}
\end{center}
\vspace{-1.5em}
  \caption{\textbf{Architecture of ZeroRF.} It parametrizes TensoRF-VM tensors with randomly-initialized deep generator networks (Sec.~\ref{sec:genarch}), with the input to the networks set to a frozen Gaussian noise on start of training. The system performs per-scene optimization using the standard volume rendering procedure with a plain rendering loss.}
\label{fig:pipeline}
\end{figure*}

\noindent \textbf{TensoRF} swapped out the initial MLP utilized in NeRF, opting for a feature volume to expedite training. It further breaks down this feature volume into factors using CANDECOMP/PARAFAC decomposition or Vector-Matrix (VM) decomposition. In our work, we mainly focus on the VM decomposition, where given a 3D tensor $\mathcal{T} \in \mathbb{R}^{I,J,K}$, it decomposes a tensor into multiple vectors
and matrices:

\begin{equation}
\centering
    \mathcal{T} = \sum_{r=1}^{R_1} v_r^1 \circ M_r^{2,3} + \sum_{r=1}^{R_2} v_r^2 \circ M_r^{1,3} + \sum_{r=1}^{R_3} v_r^3 \circ M_r^{1,2}
    \label{eq:tensorf}
\end{equation}
where $v_r^a$ are vector factors and $M_r^{b,c}$ are matrix factors.
\section{Method}

\subsection{Overview}

The ZeroRF pipeline is illustrated in Fig.~\ref{fig:pipeline}.
We use deep generator networks with a frozen standard Gaussian noise sample as input to generate the planes and vectors in the TensoRF-VM style, forming a decomposed tensorial feature volume.
The feature volume is then sampled among render rays and decoded by a multi-layer perceptron (MLP).
We employ the standard volume rendering process and a plain MSE loss.

The main idea of ZeroRF is to apply untrained deep generator networks as a parametrization of spatial feature grids.
The network can learn patterns of different scales from the sparse view observations and naturally generalize to unseen views,
without the need of further progressive upsampling tricks or explicit regularizations that typically require a lot of manual labor to tune,
as opposed to prior works for sparse view reconstruction.
Nevertheless, there are still several points of design left in the pipeline: the spatial organization, or the representation of the feature volume; the architecture of the representation generator; and the architecture of the feature decoder.
We will detail these designs in the following sections.

\subsection{Factorizing the Feature Volume}
\label{sec:volfactor}

The principle of applying deep generator networks for parametrization is universal to any grid-based representation.
The most straightforward solution is to parameterize a feature volume directly.
However, this is memory and compute inefficient as we would need a very large feature volume if we want a decent volume rendering quality.
This is not peculiar to ZeroRF; many prior arts for dense-view reconstruction actually work on this factorization. TensoRF~\cite{chen2022tensorf} uses tensorial decompositions to exploit the low-rankness of feature volumes. The triplane representation used in EG3D~\cite{chan2021eg3d} and K-planes~\cite{fridovich2023k} can be seen as a special case of TensoRF-VM representation when the vectors are constants. Dictionary Fields (DiF)~\cite{chen2023dictionary} factorizes the feature volume into multiple smaller volumes encoding different frequencies. Instant-NGP~\cite{muller2022instant} employ a multi-resolution hashmap as information in the feature volume is sparse in nature.

Among these factorizations, hashing breaks the spatial correlation between adjacent cells, so deep priors cannot be applied.
Deep generator networks can be used to parameterize all the rest three representations (TensoRF, triplane and DiF). We built generator architectures for generating 1D vectors, 2D matrices, and 3D volumes, upon which we experimented with all three factorizations. All of them work similarly and achieve better performance than previous arts, but the TensoRF-VM representation achieves slightly better performance on our test benchmarks overall. Thus, we employ the TensoRF-VM representation as our final choice of factorization, as is shown in Fig.~\ref{fig:pipeline}.

We include the comparison between different factorizations in Sec.~\ref{sec:analysis}.

\subsection{Generator Architecture}
\label{sec:genarch}

The quality of deep parametrization highly depends on the architecture.
Most generator networks to date are convolutional and attention architectures.
When designing ZeroRF, we investigated various structures including Deep Decoders (DD) \cite{heckel2018deep}, the variational autoencoder (VAE) used in Stable Diffusion (SD) \cite{podell2023sdxl}, the decoder used in Kadinsky models \cite{razzhigaev2023kandinsky}, and the SimMIM generator based on a ViT decoder \cite{xie2022simmim}. We change the 2D convolution, pooling and upsampling layers into 1D and 3D to obtain the corresponding 1D and 3D generators required in different factorizations.

These generators are originally quite large in size as they were designed to be fit on a very large dataset for generation of high-quality contents.
This results in both unnecessarily long run-times and slower convergence when it comes to fitting to a single NeRF scene.
Fortunately, we find that the performance of ZeroRF after convergence remains intact when we shrink the models in width and depth.
Thus, we keep the block composition but modify these architectures in size to boost the training speed.
Note that we only need to store the radiance field representation and not the generator during inference, so ZeroRF has zero overhead compared to its underlying factorization method during rendering.

We found that the SD VAE and its decoder part, as well as Kadinsky decoder work similarly well for novel view synthesis, followed by Deep Decoders, while the SimMIM architecture proves to be invalid as a deep prior for radiance fields. SD/Kadinsky coders are mostly convolutional architectures, with Kadinsky adding self-attention to the first two blocks. We took the (modified) SD decoder as our final choice of generator architecture as it has the least computation.
We carry out a more complete analysis on the results of using different generators in Sec.~\ref{sec:analysis}.

\subsection{Decoder Architecture}
\label{sec:decarch}

Our decoder architecture follows that of SSDNeRF \cite{chen2023single}.
We sample with linear interpolation (or bilinear, trilinear according to the dimension) from the feature grid at the point to decode, and project it with a first linear layer to get a base feature code that is shared between density and appearance decodings.
We find that sharing the feature code can help reduce floaters by coupling geometry and appearance closely.
We apply SiLU activation and invoke another linear layer for density prediction.
For color prediction, we encode the view direction with Sphere Harmonics (SH) and add its projection by a linear layer to the base feature to involve view dependence. We then apply SiLU activation and use another linear layer, similar to the density prediction, to predict RGB values. Formally, we have
\begin{align}
    \sigma_x &= \exp{\left(\Theta_\sigma(\text{SiLU}(\Theta_b(F_x)))\right)}, \\
    c_x &= \sigma\left(\Theta_c(\text{SiLU}(\Theta_b(F_x) + \Theta_d(\text{SH}(d)))\right),
\end{align}
where $F_x$ is the feature field, $\sigma(\cdot)$ is the sigmoid function, and $\Theta_\bullet$ denotes a linear weight layer.

Note that different from the decoders used in TensoRF and DiF, this decoder does not consume any positional encodings, as there would otherwise be a chance to break or degrade ZeroRF by leaking the position information outside the deep prior.

\subsection{Implementation Details}

In our experiments, we use the AdamW optimizer \cite{kingma2014adam,loshchilov2017decoupled} with $\beta_1 = 0.9, \beta_2 = 0.98$ and a weight decay of $0.2$. The learning rate starts at $0.002$ and decays to $0.001$ with a cosine schedule. We train ZeroRF for $10k$ iterations. We uniformly sample $1024$ points per ray during volume rendering, and employ occupancy pruning and occlusion culling to accelerate the process. We include figures for detailed architecture of our generator and decoder in Appendix C.

\section{Experiments}

\label{sec:exp}

\subsection{Experiment Setups}

\paragraph{Datasets and Metrics.} We evaluate our proposed method on sparse view reconstruction using NeRF-Synthetic~\cite{mildenhall2021nerf}, OpenIllumination~\cite{liu2023openillumination} and DTU~\cite{jensen2014large} datasets. We use the standard PSNR, SSIM and LPIPS~\cite{zhang2018unreasonable} metrics for evaluation.

\noindent \textbf{NeRF-Synthetic} is a synthetic dataset rendered by Blender, which contains 8 objects with various materials and geometric structures. We use 4 or 6 views as the input and evaluate the model on 200 testing views.

\noindent \textbf{OpenIllumination} is a real-world dataset captured by lightstage. We narrowed our focus to 8 objects displaying intricate geometry under a single illumination setup, extracting 4 or 6 views from the available pool of 38 training views and evaluating on 10 testing views.

\noindent \textbf{DTU} mainly focuses on forward-facing objects instead of 360$^\circ$ reconstruction, but for the sake of completeness, we include our results on DTU in Fig.~\ref{fig:dtu}. We use 3 views as the input and test the model on the rest of the views. We include more comparisons and quantitative results in Appendix B.

All the input views are selected by running KMeans~\cite{lloyd1982least,arthur2007k} on the camera translation vector and picking the views closest to cluster centroids.

\paragraph{Baselines.} We compare our ZeroRF against a few state-of-the-art few-shot NeRF methods: RegNeRF based on continuity and pertrained RealNVP~\cite{dinh2016density} regularization~\cite{niemeyer2022regnerf}, DietNeRF~\cite{jain2021putting} that uses a pretrained CLIP~\cite{radford2021learning} prior, InfoNeRF~\cite{kim2022infonerf} using entropy as regularizer, FreeNeRF~\cite{Yang2023FreeNeRF} based on frequency regularization, and FlipNeRF~\cite{seo2023flipnerf} using a spatial symmetry prior.

\subsection{Results}
\label{sec:results}

\begin{table*}[tbh]
    \centering
    \caption{Comparison with the state-of-the-art sparse view reconstruction methods on NeRF-Synthetic dataset.}
    \vspace{-0.75em}
    \small
    \begin{tabular}{lcccccccc}
    \toprule
    ~ & ~ & ~ & \multicolumn{3}{c}{4 views} & \multicolumn{3}{c}{6 views} \\
        \cmidrule{4-6}\cmidrule{7-9}
         Method & Prior & Train Time & PSNR$\uparrow$ & SSIM$\uparrow$ & LPIPS$\downarrow$ & PSNR$\uparrow$ & SSIM$\uparrow$ & LPIPS$\downarrow$ \\ \midrule
        RegNeRF~\cite{niemeyer2022regnerf} & RealNVP & $\sim$ 6 h & 9.93 & 0.419 & 0.572 & 9.82 & 0.685 & 0.580 \\ 
        DietNeRF~\cite{jain2021putting} & CLIP & $\sim$ 6 h & 10.92 & 0.557 & 0.446 & 16.92 & 0.727 & 0.267 \\ 
        InfoNeRF~\cite{kim2022infonerf} & Info. Theory & $\sim$ 3.5 h & 17.83 & 0.802 & 0.212 & 21.44 & 0.854 & 0.159 \\
        FreeNeRF~\cite{Yang2023FreeNeRF} & Frequency & $\sim$ 3 h & 18.81 & 0.808 & \cellcolor{Yellow!30}0.188 & 22.77 & \cellcolor{Yellow!30}0.865 & \cellcolor{Orange!30}0.149 \\
        FlipNeRF~\cite{seo2023flipnerf} & Symmetry & $\sim$ 3 h & 19.78 & 0.822 & 0.212  & 21.73 & 0.845 & 0.202 \\ \midrule
        Ours (1k iters) & \multirow{3}{*}{\shortstack[c]{Deep \\ Parametrization}} & 2 min & \cellcolor{Yellow!30}21.42 & \cellcolor{Yellow!30}0.843 & 0.192 & \cellcolor{Yellow!30}23.47 & 0.860 & 0.177 \\ 
        Ours (2k iters) & & 5 min & \cellcolor{Orange!30}21.85 & \cellcolor{Orange!30}0.852 & \cellcolor{Orange!30}0.170 & \cellcolor{Orange!30}24.29 & \cellcolor{Orange!30}0.875 & \cellcolor{Yellow!30}0.153 \\ 
        Ours (Full) & & $\sim$ 25 min & \cellcolor{Red!30}\textbf{21.94} & \cellcolor{Red!30}\textbf{0.856} & \cellcolor{Red!30}\textbf{0.139} & \cellcolor{Red!30}\textbf{24.73} & \cellcolor{Red!30}\textbf{0.889} & \cellcolor{Red!30}\textbf{0.113} \\ 
    \bottomrule
    \end{tabular}
\label{tab:nerfsyn}
\end{table*}

\begin{figure*}[tbh]
\begin{center}
   \includegraphics[width=0.95\linewidth]{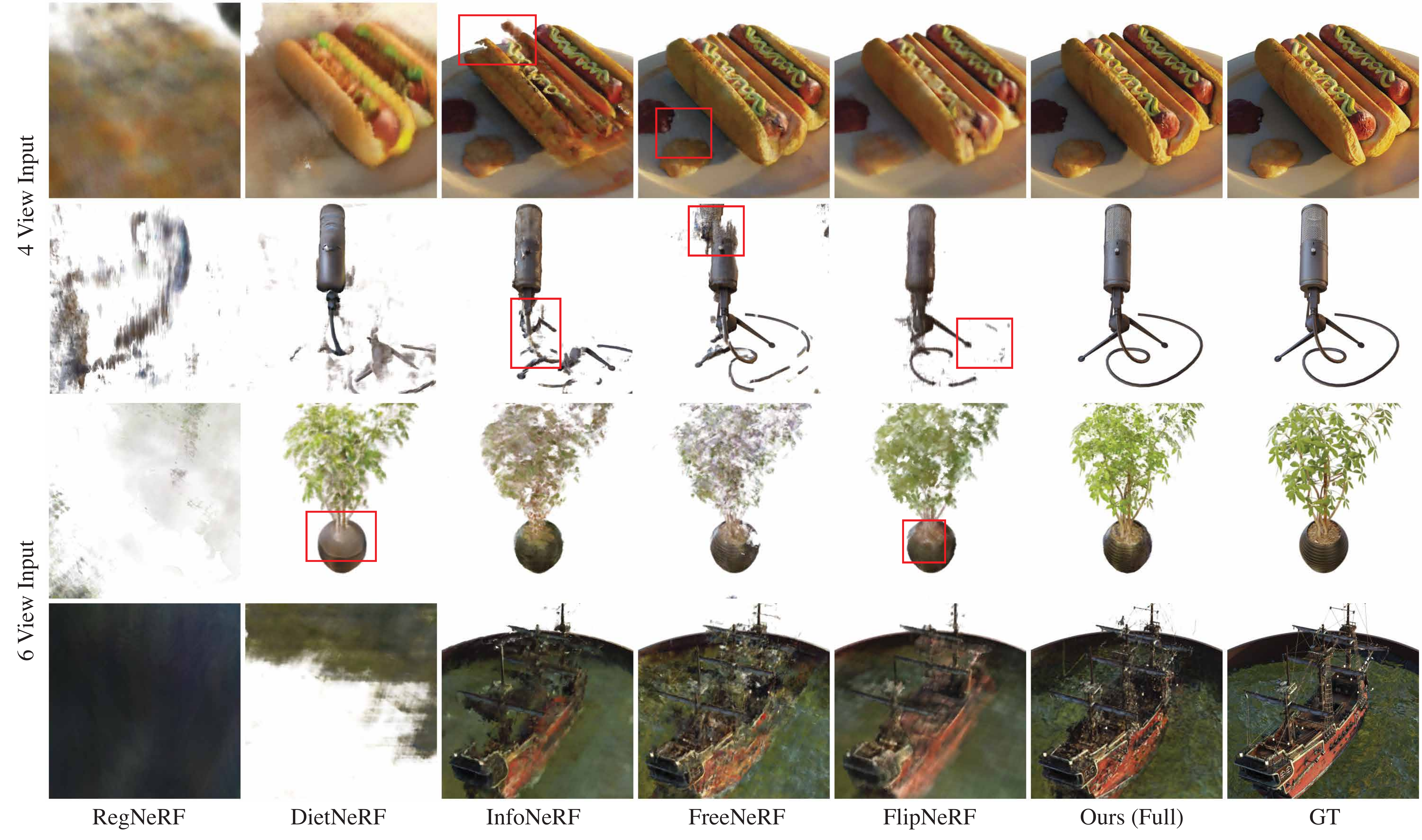}
\end{center}
\vspace{-1.5em}
  \caption{\textbf{Qualitative comparison between ZeroRF and previous works on NeRF-Synthetic dataset.} The top two rows (Hotdog and Mic) are reconstruction results from 4 views and the bottom two rows (Ficus and Ship) are reconstruction results from 6 views. ZeroRF results have the best visual quality, and is free of walls or floaters.}
\label{fig:nerfsyn}
\end{figure*}

\begin{figure*}[tbh]
\begin{center}
   \includegraphics[width=0.95\linewidth]{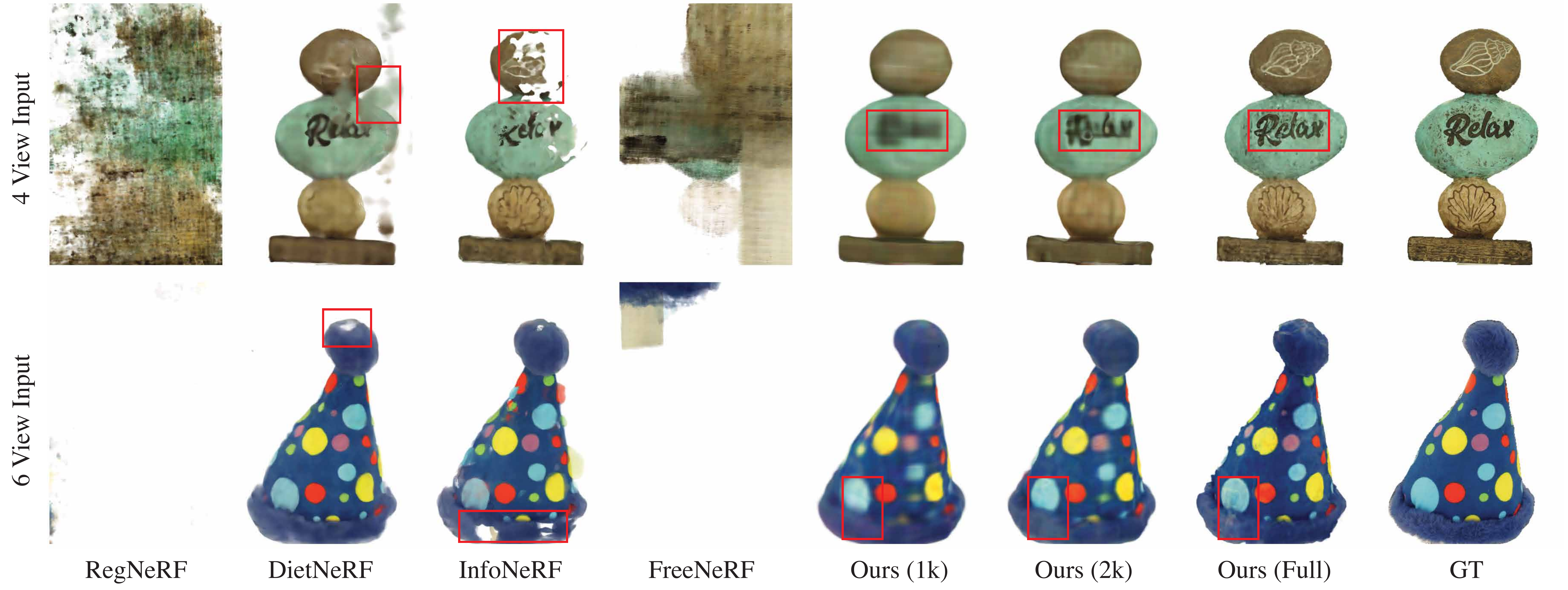}
\end{center}
\vspace{-1.5em}
  \caption{Qualitative comparison between ZeroRF and previous works on OpenIllumination dataset.}
\label{fig:oppo}
\end{figure*}

\begin{figure}[tbh]
\begin{center}
   \includegraphics[width=0.95\linewidth]{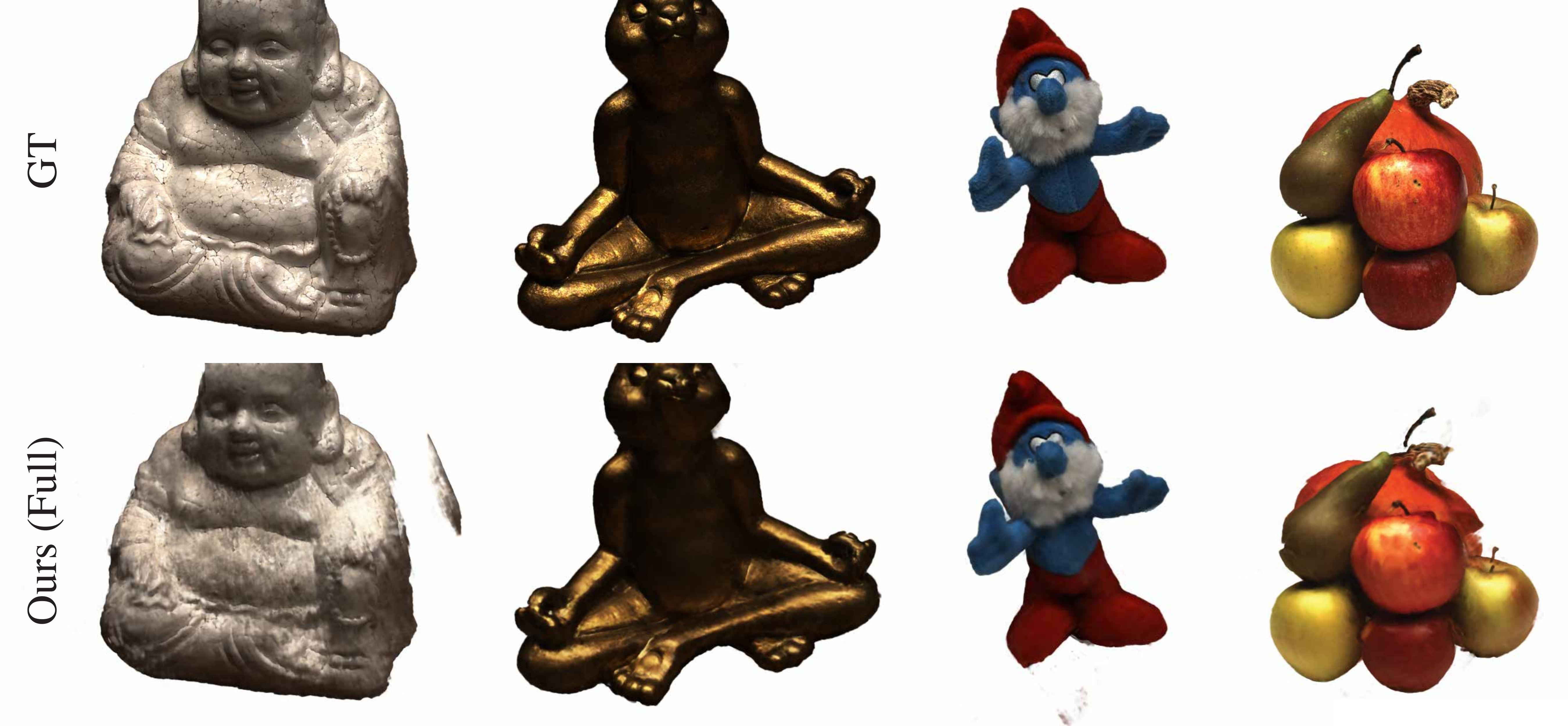}
\end{center}
\vspace{-1.5em}
  \caption{\textbf{Results of our method on DTU with only 3 views as input.} See Appendix B for more details and comparisons.}
\label{fig:dtu}
\end{figure}

The quantitative results for NeRF-Synthetic and OpenIllumination are presented in Tab.~\ref{tab:nerfsyn} and Tab.~\ref{tab:oppo}, respectively. Across both the 4-view and 6-view experiments, our approach consistently outperforms all other methods, as evidenced by superior PSNR, SSIM, and LPIPS scores. Moreover, our method achieves these results in significantly less time. Even with only 2 minutes of training, ZeroRF remains superior or competitive to the best baselines.

\begin{table}[tbh]
\small
    \centering
\caption{Comparison with the state-of-the-art sparse view reconstruction methods on OpenIllumination.}
\vspace{-0.75em}
    \begin{tabular}{lcccc}
    \toprule
         ~& Train Time & PSNR$\uparrow$ & SSIM$\uparrow$ & LPIPS$\downarrow$ \\ \midrule
        RegNeRF~\cite{niemeyer2022regnerf} & $\sim$ 6 h & 14.08 & 0.859 & 0.303 \\ 
        DietNeRF~\cite{jain2021putting} & $\sim$ 6 h & 24.20 & \cellcolor{Orange!30}0.940 & 0.095 \\ 
        InfoNeRF~\cite{kim2022infonerf} & $\sim$ 3.5 h & 22.28 & 0.935 & \cellcolor{Orange!30}0.090 \\
        FreeNeRF~\cite{Yang2023FreeNeRF} & $\sim$ 3 h & 11.47 & 0.813 & 0.351 \\\midrule
        Ours (1k iters) & 2 min & \cellcolor{Orange!30}26.26 & \cellcolor{Yellow!30}0.939 & 0.097 \\ 
        Ours (2k iters) & 5 min & \cellcolor{Red!30}26.49 & \cellcolor{Red!30}0.941 & \cellcolor{Orange!30}0.090 \\
        Ours (Full) & $\sim$ 25 min & \cellcolor{Yellow!30}26.15 & \cellcolor{Yellow!30}0.939 & \cellcolor{Red!30}0.080 \\
    \bottomrule
    \end{tabular}
\label{tab:oppo}
\end{table}

Visual comparisons between ZeroRF and baseline methods are illustrated in Fig.~\ref{fig:nerfsyn} and Fig.~\ref{fig:oppo}. Most of the baseline models exhibit noticeable flaws of varying degrees, including floaters and apparent color shifts in synthesis results (highlighted within red boxes in the figure). For pretrained priors, the RegNeRF prior model was not trained on wide-baseline images, and fails to reconstruct objects under 360$^\circ$ settings; DietNeRF using CLIP as a prior model interestingly works better on real images than on synthetic images, which is consistent with CLIP's pretraining data distribution. For non-pretrained models, InfoNeRF and FreeNeRF applying information-theoretical and frequency regularizers fail to represent intricate structures like Ficus leaves. Notably, FreeNeRF and FlipNeRF perform relatively well on NeRF-Synthetic, but fail catastrophically on OpenIllumination. This shows that handcrafted priors are not robust to setting changes. FlipNeRF fails with numerical instabilities during training on OpenIllumination, which is also observed by Wang on their own data \cite{wangpan}. ZeroRF shows
the best visual quality and robustness across diverse datasets and is free of floaters or unrealistic color shifts on all scenes.

We refer the readers to Appendix A for more detailed results on the two benchmarks.

\subsection{Analysis}
\label{sec:analysis}

\paragraph{Effect of the Number of Training Views.}
We designed experiments to show the benefit of our proposed method with the number of input views.
We plot the results in Fig.~\ref{fig:views}.
ZeroRF has a significant advantage to the base TensoRF representation on sparse views (3 to 8).
When the views become denser, ZeroRF remains competitive, though by a smaller margin.

\begin{figure}[ht]
\begin{center}
   \includegraphics[width=0.85\linewidth]{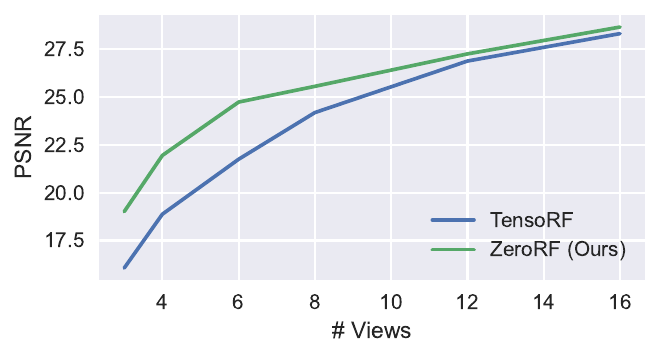}
\end{center}
\vspace{-2em}
  \caption{PSNR of ZeroRF versus vanilla TensoRF on NeRF-Synthetic dataset.}
\label{fig:views}
\end{figure}

\begin{table}[tbh]
\centering
\caption{\textbf{Applying our prior to different grid-based representations.} ZeroRF parametrization consistently enhances the models to better generalize to unseen views. Results are from the 6-view NeRF-Synthetic setting.}
\label{tab:factorization}
\vspace{-0.75em}
\small
\begin{tabular}{lcccc} 
\toprule
Grid Repr.                  & Our Prior & PSNR$\uparrow$ & SSIM$\uparrow$ & LPIPS$\downarrow$  \\ 
\midrule
\multirow{2}{*}{Triplane~\cite{chan2021eg3d}} &     \xmark        & 20.55          & 0.822          & 0.201              \\
                            &     \cmark       & \textbf{24.38} & \textbf{0.887} & \textbf{0.117}     \\ 
\midrule
\multirow{2}{*}{DiF~\cite{chen2023dictionary}}       &      \xmark       & 17.45          & 0.775          & 0.258              \\
                            &      \cmark      & \textbf{23.56} & \textbf{0.878} & \textbf{0.125}     \\ 
\midrule
\multirow{2}{*}{TensoRF~\cite{chen2022tensorf}} &     \xmark        & 21.73          & 0.844          & 0.169              \\
                            &     \cmark       & \textbf{24.73} & \textbf{0.889} & \textbf{0.113}     \\
\bottomrule
\end{tabular}
\end{table}
\paragraph{Feature Volume Factorization Choices.} We apply ZeroRF generators to Triplane, TensoRF and DiF and compare the performance of resulting parametrizations on the NeRF-Synthetic dataset (6-view setting). The results are shown in Tab.~\ref{tab:factorization}.
The inclusion of generators consistently improves upon base representations, and they all achieve state-of-the-art performance.
This shows that the principles of using a deep parametrization is generally applicable to grid-based representations (also see Sec.~\ref{sec:volfactor}).
Among the factorization methods, TensoRF with our prior performs the best, so we chose TensoRF as our final feature volume factorization.

\setlength{\tabcolsep}{5pt}
\begin{table}[tbh]
\small
    \centering
    \caption{\textbf{Ablation study on VM generator architecture.} Results are from the 6-view NeRF-Synthetic setting. `Up' in the table refers to bilinear upsampling.}
    \small
    \vspace{-0.75em}
    \begin{tabular}{lcccc}
    \toprule
        Generator & Architecture & PSNR$\uparrow$ & SSIM$\uparrow$ & LPIPS$\downarrow$ \\ 
        \midrule
        No Generator         & - & 21.73 & 0.844 & 0.169 \\ 
        MLP on Grid          & MLP & 20.69 & 0.827 & 0.214 \\
        SimMIM~\cite{xie2022simmim} & Attn. (ViT) & 21.40 & 0.839 & 0.156 \\
        SD VAE~\cite{podell2023sdxl} & Conv. & 24.68 & 0.890 & \cellcolor{Yellow!0}0.116 \\
        Kadinsky~\cite{razzhigaev2023kandinsky} & Attn. + Conv. & \cellcolor{Orange!0}24.69 & \cellcolor{Red!0}0.889 & \cellcolor{Red!0}0.112 \\
        DD~\cite{heckel2018deep} & Lin. + Up & 23.93 & 0.876 & 0.148 \\
        \midrule
        SD Decoder         & Conv. & \cellcolor{Red!0}24.73 & \cellcolor{Red!0}0.889 & \cellcolor{Orange!0}0.113 \\
        \quad 2$\times$ layers & Conv. & \cellcolor{Orange!0}24.69 & \cellcolor{Red!0}0.889 & 0.122 \\
        \quad No Noise         & Conv. & 12.21 & 0.785 & 0.375 \\
    \bottomrule
    \end{tabular}
\label{tab:priorabl}
\end{table}
\begin{figure}[ht]
\begin{center}
   \includegraphics[width=\linewidth]{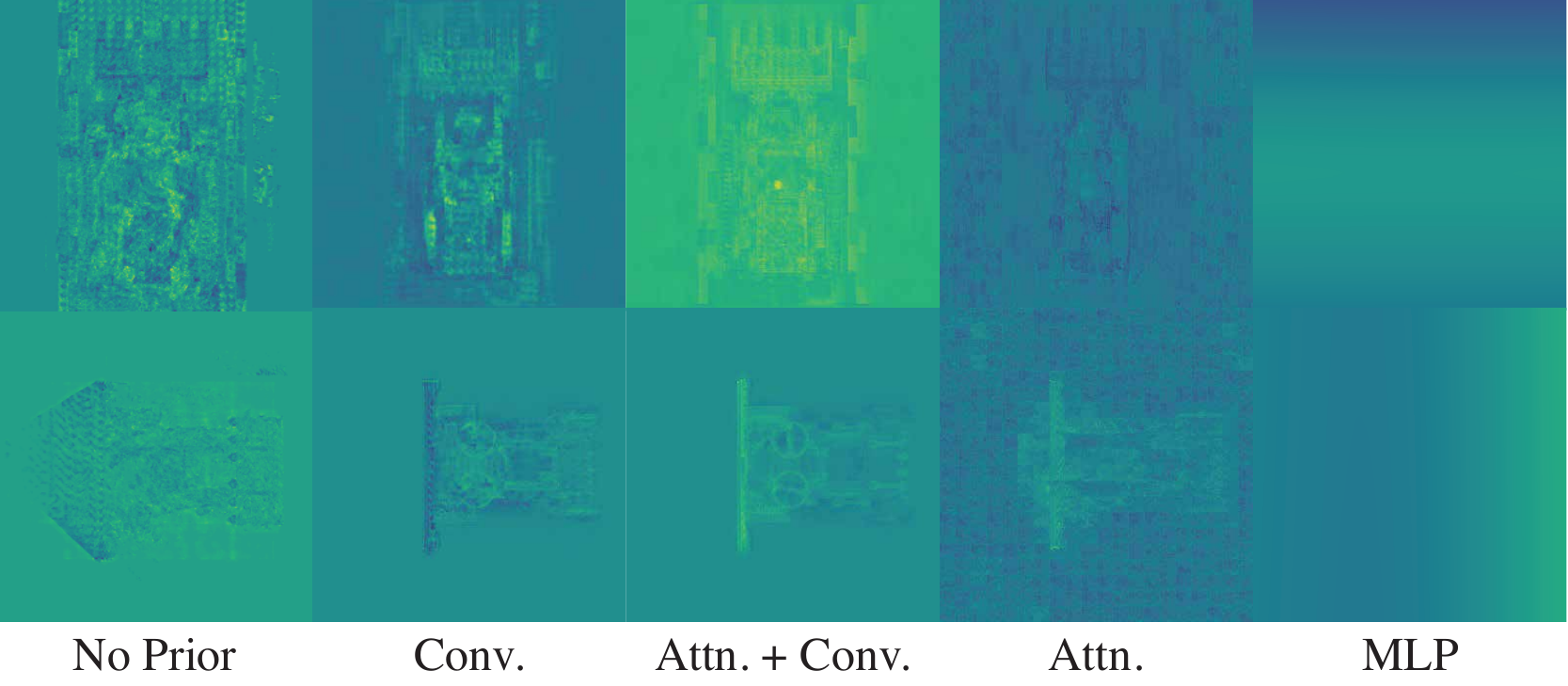}
\end{center}
\vspace{-1.5em}
  \caption{\textbf{Visualization of plane features from different generators.} Different architecture impose different priors on features.}
\label{fig:features}
\end{figure}
\paragraph{Generator Architecture.} We applied different generator architectures introduced in Sec.~\ref{sec:genarch} upon the TensoRF factorization and compared their performance in Tab.~\ref{tab:priorabl}. To further investigate the effect of different priors, we visualize one channel of the plane features with different generators in Fig.~\ref{fig:features}. Without any prior and directly optimizing the planes, the features are noisy with high-frequency glitches and visible view boundary lines all across. In contrast, the SD Decoder and Kadinsky models produce clean and well-post features. The fully attentional ViT decoder of SimMIM works with patches, and we can see visible blocky artifacts. MLP assumes a very smooth transition over the grid and thus is unable to represent scene content faithfully. Overall, the convolutional architectures produce features that align the best with the scene.

ZeroRF is robust to over-parametrization of the networks. The results are similar if we scale the decoder with 2x more layers (the second last row in Tab.~\ref{tab:priorabl}).

\paragraph{Importance of the Noise.} The input noise is the key to our prior. Swapping it with a trainable feature initialized with zeros breaks the system completely (the last row in Tab.~\ref{tab:priorabl}). We do not observe performance improvements if we unfreeze the noise -- as the learning rate would be small compared to the scale of the noise, the structure of the noise is kept unchanged throughout training. But it introduces extra overhead and slows down the convergence. Thus, we keep the noise frozen during training.

\begin{figure}[tbh]
\begin{center}
   \includegraphics[width=\linewidth]{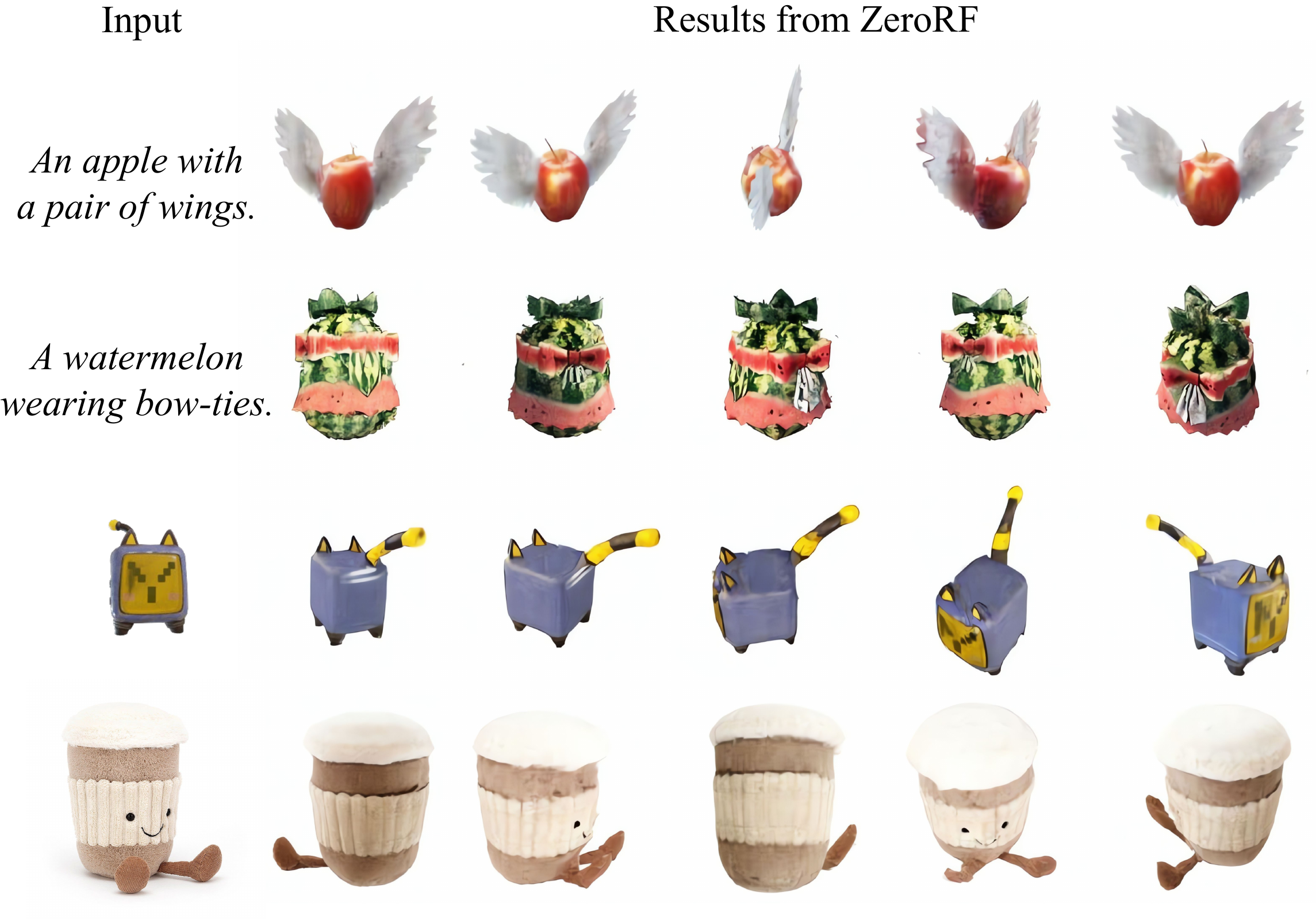}
\end{center}
\vspace{-1.5em}
  \caption{\textbf{Text-to-3D and Image-to-3D generation results with ZeroRF.} ZeroRF can naturally handle model-generated multi-view images, and reconstruct $360^\circ$ views from the sparse view generations with high quality in 30 seconds.}
\label{fig:gen3d}
\end{figure}

\begin{figure}[tbh]
\begin{center}
   \includegraphics[width=\linewidth]{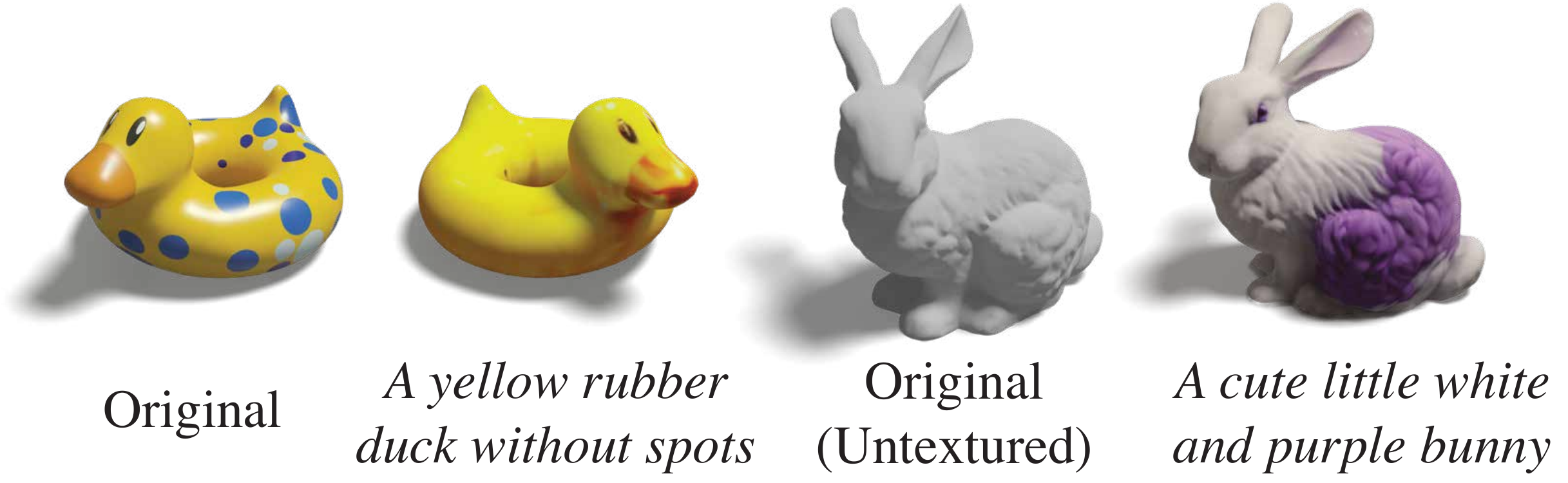}
\end{center}
\vspace{-1.5em}
  \caption{\textbf{Texture generation with ZeroRF.} ZeroRF can be used to apply new appearance to a given geometry, with the assistance of language-based image editing models.}
\label{fig:texedit}
\end{figure}

\section{Applications}
\label{sec:app}

\paragraph{Text to 3D and Image to 3D.} Given the powerful sparse-view reconstruction capability of ZeroRF, a straightforward idea is to use an existing model to perform consistent multi-view generation, and apply ZeroRF to lift the sparse view into 3D. In this example, for image-to-3D we employ Zero123++~\cite{shi2023zero123++} to lift single image input into 6-view images and directly fit a ZeroRF on the generated images. For text-to-3D, we first invoke SDXL \cite{podell2023sdxl} to generate an image from the text, and apply the image-to-3D procedure described before. As shown in Fig.~\ref{fig:gen3d}, ZeroRF is able to produce faithful high-quality reconstructions from generated multi-view images. Fitting the ZeroRF only costs 30 seconds on a single A100 GPU.

\paragraph{Mesh Texturing and Texture Editing.} ZeroRF can also be utilized to reconstruct the appearance with a frozen provided geometry. To do this, we render 4 images of the mesh from random views, tile them into one large image and apply Instruct-Pix2Pix \cite{brooks2023instructpix2pix} to edit the images according to a text prompt. We then fit a ZeroRF on the four images, and bake the color values back to the mesh surface. In this case, fitting the ZeroRF only requires 20 seconds. Fig.~\ref{fig:texedit} (and also the rightmost column in Fig.~1) demonstrate results of texture editing on the Bob mesh (Fig.~\ref{fig:texedit} Left) and mesh texturing on Stanford Bunny (Fig.~\ref{fig:texedit} Right).

\section{Conclusion and Future Work}

In this work we present ZeroRF, a novel method for fast and high quality sparse view 360$^{\circ}$ reconstruction.
Based on a deep parametrization technique, it can be applied on various factorized grid-based radiance fields, achieving state-of-the-art performance for sparse view 360$^{\circ}$ reconstruction without the need of designing any specific regularizations or incorporating any pretraining priors.

One possible future work would be extending ZeroRF to unbounded scenes; we discuss more limitations and future work in Appendix D.

\section*{Acknowledgement}
This work is supported in part by gifts from Qualcomm.

\fi

{
    \small
    \bibliographystyle{ieeenat_fullname}
    \bibliography{main}
}

\if\supp1
\appendix
\section{More Results on NeRF-Synthetic and OpenIllumination}

Here we show more complete results including metrics and visualization views on all scenes in NeRF-Synthetic and OpenIllumination. See Tab.~\ref{tab:nerf6vfull}, \ref{tab:nerf4vfull}, \ref{tab:oppo6vfull}, \ref{tab:oppo4vfull} and Fig.~\ref{fig:nerfsyn_supp}, \ref{fig:oppo_supp}.

\begin{table*}[tbh]
    \caption{Comparison of per-scene metrics of NeRF-Synthetic 6 view settings.}
    \label{tab:nerf6vfull}
    \vspace{-0.75em}
    \centering
    \small
    \begin{tabularx}{0.98\textwidth}{l c *{9}{Y}}
\toprule
 &  & chair & drums & ficus & hotdog & lego & materials & mic & ship & mean \\
\midrule
 & PSNR & 13.16 & 9.44 & 12.34 & 9.61 & 9.07 & 7.40 & 11.28 & 6.19 & 9.81 \\
RegNeRF & SSIM & 0.580 & 0.280 & 0.512 & 0.470 & 0.413 & 0.258 & 0.463 & 0.285 & 0.407 \\
 & LPIPS & 0.510 & 0.609 & 0.543 & 0.569 & 0.600 & 0.634 & 0.523 & 0.651 & 0.580 \\
\midrule
 & PSNR & \cellcolor{Yellow!30}25.31 & \cellcolor{Yellow!30}18.66 & 19.81 & \cellcolor{Orange!30}27.73 & 21.29 & \cellcolor{Yellow!30}20.79 & 20.38 & 19.84 & \cellcolor{Yellow!30}21.73 \\
FlipNeRF & SSIM & 0.887 & 0.815 & 0.844 & \cellcolor{Yellow!30}0.925 & 0.820 & \cellcolor{Yellow!30}0.839 & 0.887 & \cellcolor{Yellow!30}0.746 & 0.845 \\
 & LPIPS & \cellcolor{Yellow!30}0.080 & 0.239 & 0.144 & 0.173 & 0.207 & 0.242 & 0.172 & 0.361 & 0.202 \\
\midrule
 & PSNR & 25.21 & \cellcolor{Orange!30}19.81 & \cellcolor{Yellow!30}20.18 & 9.37 & 20.31 & 7.78 & \cellcolor{Orange!30}25.85 & 6.79 & 16.91 \\
DietNeRF & SSIM & 0.887 & \cellcolor{Orange!30}0.838 & \cellcolor{Yellow!30}0.852 & 0.644 & 0.805 & 0.447 & \cellcolor{Yellow!30}0.941 & 0.405 & 0.727 \\
 & LPIPS & 0.112 & \cellcolor{Orange!30}0.133 & \cellcolor{Orange!30}0.127 & 0.442 & 0.174 & 0.493 & \cellcolor{Yellow!30}0.072 & 0.582 & 0.267 \\
\midrule
 & PSNR & 24.87 & 18.39 & \cellcolor{Orange!30}20.59 & 23.61 & \cellcolor{Yellow!30}21.92 & 20.42 & 20.84 & \cellcolor{Orange!30}20.86 & 21.44 \\
InfoNeRF & SSIM & \cellcolor{Yellow!30}0.892 & 0.824 & \cellcolor{Orange!30}0.859 & 0.903 & \cellcolor{Yellow!30}0.854 & 0.838 & 0.904 & \cellcolor{Red!30}0.758 & \cellcolor{Yellow!30}0.854 \\
 & LPIPS & 0.111 & 0.190 & \cellcolor{Yellow!30}0.138 & \cellcolor{Yellow!30}0.133 & \cellcolor{Yellow!30}0.150 & \cellcolor{Orange!30}0.159 & 0.119 & \cellcolor{Orange!30}0.274 & \cellcolor{Yellow!30}0.159 \\
\midrule
 & PSNR & \cellcolor{Orange!30}26.57 & 18.16 & 18.46 & \cellcolor{Yellow!30}27.18 & \cellcolor{Orange!30}24.32 & \cellcolor{Red!30}21.63 & \cellcolor{Yellow!30}25.64 & \cellcolor{Yellow!30}20.23 & \cellcolor{Orange!30}22.77 \\
FreeNeRF & SSIM & \cellcolor{Orange!30}0.916 & \cellcolor{Yellow!30}0.827 & 0.840 & \cellcolor{Orange!30}0.929 & \cellcolor{Orange!30}0.887 & \cellcolor{Red!30}0.853 & \cellcolor{Orange!30}0.942 & 0.729 & \cellcolor{Orange!30}0.865 \\    
 & LPIPS & \cellcolor{Red!30}0.071 & \cellcolor{Yellow!30}0.176 & 0.161 & \cellcolor{Orange!30}0.096 & \cellcolor{Orange!30}0.132 & \cellcolor{Yellow!30}0.202 & \cellcolor{Orange!30}0.066 & \cellcolor{Yellow!30}0.290 & \cellcolor{Orange!30}0.149 \\
\midrule
 & PSNR & \cellcolor{Red!30}27.62 & \cellcolor{Red!30}20.88 & \cellcolor{Red!30}22.21 & \cellcolor{Red!30}29.93 & \cellcolor{Red!30}26.26 & \cellcolor{Orange!30}21.41 & \cellcolor{Red!30}27.40 & \cellcolor{Red!30}22.13 & \cellcolor{Red!30}24.73 \\
Ours & SSIM & \cellcolor{Red!30}0.926 & \cellcolor{Red!30}0.869 & \cellcolor{Red!30}0.898 & \cellcolor{Red!30}0.949 & \cellcolor{Red!30}0.913 & \cellcolor{Orange!30}0.849 & \cellcolor{Red!30}0.954 & \cellcolor{Orange!30}0.756 & \cellcolor{Red!30}0.889 \\
 & LPIPS & \cellcolor{Orange!30}0.074 & \cellcolor{Red!30}0.131 & \cellcolor{Red!30}0.100 & \cellcolor{Red!30}0.075 & \cellcolor{Red!30}0.085 & \cellcolor{Red!30}0.132 & \cellcolor{Red!30}0.050 & \cellcolor{Red!30}0.256 & \cellcolor{Red!30}0.113 \\
 \bottomrule
    \end{tabularx}
\end{table*}

\begin{table*}[tbh]
    \caption{Comparison of per-scene metrics of NeRF-Synthetic 4 view settings.}
    \label{tab:nerf4vfull}
    \vspace{-0.75em}
    \centering
    \small
    \begin{tabularx}{0.98\textwidth}{l c *{9}{Y}}
\toprule
 &  & chair & drums & ficus & hotdog & lego & materials & mic & ship & mean \\
\midrule
 & PSNR & 13.12 & 9.75 & 11.78 & 9.16 & 8.64 & 7.91 & 13.10 & 5.98 & 9.93 \\
RegNeRF & SSIM & 0.581 & 0.304 & 0.422 & 0.475 & 0.364 & 0.254 & 0.696 & 0.258 & 0.419 \\
 & LPIPS & 0.507 & 0.615 & 0.594 & 0.565 & 0.639 & 0.621 & 0.353 & 0.683 & 0.572 \\
\midrule
 & PSNR & 19.89 & \cellcolor{Orange!30}16.53 & \cellcolor{Yellow!30}18.76 & \cellcolor{Orange!30}26.26 & \cellcolor{Yellow!30}19.96 & \cellcolor{Orange!30}20.71 & \cellcolor{Orange!30}17.99 & \cellcolor{Yellow!30}18.15 & \cellcolor{Orange!30}19.78 \\
FlipNeRF & SSIM & 0.828 & \cellcolor{Orange!30}0.771 & \cellcolor{Yellow!30}0.836 & \cellcolor{Orange!30}0.918 & \cellcolor{Yellow!30}0.801 & 0.844 & \cellcolor{Orange!30}0.858 & \cellcolor{Red!30}0.715 & \cellcolor{Orange!30}0.822 \\    
 & LPIPS & \cellcolor{Yellow!30}0.130 & \cellcolor{Yellow!30}0.281 & \cellcolor{Yellow!30}0.151 & \cellcolor{Yellow!30}0.170 & \cellcolor{Yellow!30}0.209 & 0.200 & \cellcolor{Orange!30}0.182 & 0.374 & \cellcolor{Yellow!30}0.212 \\        
\midrule
 & PSNR & 17.47 & 12.96 & 9.50 & 12.33 & 7.87 & 6.19 & 14.81 & 6.21 & 10.92 \\
DietNeRF & SSIM & 0.775 & 0.650 & 0.451 & 0.658 & 0.397 & 0.363 & 0.773 & 0.389 & 0.557 \\
 & LPIPS & 0.264 & 0.333 & 0.518 & 0.418 & 0.587 & 0.545 & 0.286 & 0.616 & 0.446 \\
\midrule
 & PSNR & \cellcolor{Yellow!30}20.02 & 12.13 & \cellcolor{Orange!30}19.47 & 18.92 & 17.77 & 20.38 & \cellcolor{Yellow!30}15.79 & \cellcolor{Orange!30}18.18 & 17.83 \\
InfoNeRF & SSIM & \cellcolor{Yellow!30}0.841 & 0.686 & \cellcolor{Orange!30}0.849 & 0.864 & 0.770 & \cellcolor{Orange!30}0.850 & \cellcolor{Yellow!30}0.845 & \cellcolor{Orange!30}0.713 & 0.802 \\
 & LPIPS & 0.164 & 0.344 & 0.153 & 0.179 & 0.221 & \cellcolor{Orange!30}0.142 & \cellcolor{Yellow!30}0.184 & \cellcolor{Orange!30}0.310 & \cellcolor{Yellow!30}0.212 \\
\midrule
 & PSNR & \cellcolor{Orange!30}20.22 & \cellcolor{Yellow!30}14.99 & 17.35 & \cellcolor{Yellow!30}23.58 & \cellcolor{Orange!30}20.43 & \cellcolor{Red!30}21.36 & 15.05 & 17.52 & \cellcolor{Yellow!30}18.81 \\
FreeNeRF & SSIM & \cellcolor{Orange!30}0.843 & \cellcolor{Yellow!30}0.746 & 0.809 & \cellcolor{Yellow!30}0.899 & \cellcolor{Orange!30}0.818 & \cellcolor{Red!30}0.857 & 0.802 & 0.687 & \cellcolor{Yellow!30}0.808 \\
 & LPIPS & \cellcolor{Orange!30}0.109 & \cellcolor{Orange!30}0.280 & \cellcolor{Orange!30}0.144 & \cellcolor{Orange!30}0.108 & \cellcolor{Orange!30}0.156 & \cellcolor{Yellow!30}0.174 & 0.218 & \cellcolor{Yellow!30}0.318 & \cellcolor{Orange!30}0.188 \\
\midrule
 & PSNR & \cellcolor{Red!30}23.04 & \cellcolor{Red!30}16.91 & \cellcolor{Red!30}20.12 & \cellcolor{Red!30}29.11 & \cellcolor{Red!30}22.11 & \cellcolor{Yellow!30}20.50 & \cellcolor{Red!30}24.76 & \cellcolor{Red!30}19.01 & \cellcolor{Red!30}21.94 \\
Ours & SSIM & \cellcolor{Red!30}0.880 & \cellcolor{Red!30}0.791 & \cellcolor{Red!30}0.866 & \cellcolor{Red!30}0.944 & \cellcolor{Red!30}0.868 & \cellcolor{Yellow!30}0.848 & \cellcolor{Red!30}0.944 & \cellcolor{Yellow!30}0.707 & \cellcolor{Red!30}0.856 \\
 & LPIPS & \cellcolor{Red!30}0.107 & \cellcolor{Red!30}0.206 & \cellcolor{Red!30}0.120 & \cellcolor{Red!30}0.088 & \cellcolor{Red!30}0.122 & \cellcolor{Red!30}0.129 & \cellcolor{Red!30}0.056 & \cellcolor{Red!30}0.283 & \cellcolor{Red!30}0.139 \\
 \bottomrule
    \end{tabularx}
\end{table*}

\begin{table*}[tbh]
    \caption{Comparison of per-scene metrics of OpenIllumination 6 view settings. We employ early-stopping by error on a validation view.}
    \label{tab:oppo6vfull}
    \vspace{-0.75em}
    \centering
    \small
    \begin{tabularx}{0.98\textwidth}{l c *{9}{Y}}
\toprule
 &  & stone & pumpkin & toy & potato & pine & shroom & cow & cake & mean \\
\midrule
 & PSNR & 13.80 & 13.58 & 13.54 & 13.92 & 11.87 & 13.22 & 13.07 & \cellcolor{Yellow!30}19.66 & 14.08 \\
RegNeRF & SSIM & 0.848 & 0.848 & 0.884 & 0.854 & 0.807 & 0.863 & 0.807 & 0.958 & 0.859 \\
 & LPIPS & 0.288 & 0.350 & 0.237 & 0.348 & 0.337 & 0.329 & 0.405 & 0.128 & 0.303 \\
\midrule
 & PSNR & \cellcolor{Orange!30}24.87 & \cellcolor{Yellow!30}24.80 & \cellcolor{Yellow!30}25.37 & \cellcolor{Orange!30}25.63 & \cellcolor{Yellow!30}18.16 & \cellcolor{Orange!30}23.71 & \cellcolor{Yellow!30}21.50 & \cellcolor{Orange!30}29.58 & \cellcolor{Orange!30}24.20 \\
DietNeRF & SSIM & \cellcolor{Orange!30}0.921 & \cellcolor{Red!30}0.966 & \cellcolor{Yellow!30}0.944 & \cellcolor{Red!30}0.955 & \cellcolor{Yellow!30}0.902 & \cellcolor{Red!30}0.930 & \cellcolor{Red!30}0.930 & \cellcolor{Red!30}0.973 & \cellcolor{Orange!30}0.940 \\
 & LPIPS & \cellcolor{Orange!30}0.085 & \cellcolor{Yellow!30}0.073 & \cellcolor{Yellow!30}0.086 & \cellcolor{Yellow!30}0.087 & \cellcolor{Yellow!30}0.119 & \cellcolor{Yellow!30}0.119 & \cellcolor{Yellow!30}0.133 & \cellcolor{Orange!30}0.059 & \cellcolor{Yellow!30}0.095 \\
\midrule
 & PSNR & \cellcolor{Yellow!30}14.37 & \cellcolor{Orange!30}26.02 & \cellcolor{Orange!30}25.91 & \cellcolor{Yellow!30}25.55 & \cellcolor{Orange!30}21.71 & \cellcolor{Yellow!30}22.99 & \cellcolor{Orange!30}22.04 & 19.60 & \cellcolor{Yellow!30}22.27 \\
InfoNeRF & SSIM & \cellcolor{Yellow!30}0.910 & \cellcolor{Yellow!30}0.960 & \cellcolor{Red!30}0.952 & \cellcolor{Yellow!30}0.946 & \cellcolor{Orange!30}0.917 & \cellcolor{Yellow!30}0.914 & \cellcolor{Yellow!30}0.915 & \cellcolor{Yellow!30}0.962 & \cellcolor{Yellow!30}0.935 \\
 & LPIPS & \cellcolor{Yellow!30}0.106 & \cellcolor{Red!30}0.057 & \cellcolor{Red!30}0.058 & \cellcolor{Red!30}0.078 & \cellcolor{Red!30}0.086 & \cellcolor{Orange!30}0.116 & \cellcolor{Orange!30}0.126 & \cellcolor{Yellow!30}0.094 & \cellcolor{Orange!30}0.090 \\
\midrule
 & PSNR & 11.62 & 11.71 & 10.65 & 11.35 & 8.85 & 10.12 & 11.09 & 16.33 & 11.46 \\
FreeNeRF & SSIM & 0.791 & 0.864 & 0.814 & 0.832 & 0.753 & 0.764 & 0.784 & 0.900 & 0.813 \\
 & LPIPS & 0.236 & 0.293 & 0.346 & 0.397 & 0.328 & 0.505 & 0.442 & 0.265 & 0.351 \\
\midrule
 & PSNR & \cellcolor{Red!30}26.30 & \cellcolor{Red!30}27.87 & \cellcolor{Red!30}27.28 & \cellcolor{Red!30}27.26 & \cellcolor{Red!30}22.26 & \cellcolor{Red!30}26.34 & \cellcolor{Red!30}23.74 & \cellcolor{Red!30}31.00 & \cellcolor{Red!30}26.49 \\
Ours & SSIM & \cellcolor{Red!30}0.929 & \cellcolor{Red!30}0.966 & \cellcolor{Orange!30}0.950 & \cellcolor{Orange!30}0.951 & \cellcolor{Red!30}0.918 & \cellcolor{Orange!30}0.928 & \cellcolor{Orange!30}0.921 & \cellcolor{Orange!30}0.969 & \cellcolor{Red!30}0.941 \\
 & LPIPS & \cellcolor{Red!30}0.063 & \cellcolor{Orange!30}0.064 & \cellcolor{Orange!30}0.062 & \cellcolor{Orange!30}0.084 & \cellcolor{Orange!30}0.088 & \cellcolor{Red!30}0.106 & \cellcolor{Red!30}0.118 & \cellcolor{Red!30}0.052 & \cellcolor{Red!30}0.080 \\
 \bottomrule
    \end{tabularx}
\end{table*}

\begin{table*}[tbh]
    \caption{Comparison of per-scene metrics of OpenIllumination 4 view settings. We employ early-stopping by error on a validation view.}
    \label{tab:oppo4vfull}
    \vspace{-0.75em}
    \centering
    \small
    \begin{tabularx}{0.98\textwidth}{l c *{9}{Y}}
\toprule
 &  & stone & pumpkin & toy & potato & pine & shroom & cow & cake & mean \\
\midrule
 & PSNR & 10.26 & 11.74 & 10.04 & 11.63 & 9.37 & 10.66 & 11.99 & 17.21 & 11.62 \\
RegNeRF & SSIM & 0.602 & 0.749 & 0.637 & 0.719 & 0.571 & 0.658 & 0.748 & 0.868 & 0.694 \\
 & LPIPS & 0.483 & 0.465 & 0.476 & 0.505 & 0.486 & 0.551 & 0.460 & 0.359 & 0.473 \\
\midrule
 & PSNR & \cellcolor{Yellow!30}24.05 & \cellcolor{Red!30}26.54 & \cellcolor{Red!30}24.98 & \cellcolor{Orange!30}23.00 & \cellcolor{Red!30}20.94 & \cellcolor{Orange!30}19.91 & \cellcolor{Orange!30}16.30 & \cellcolor{Orange!30}28.97 & \cellcolor{Orange!30}23.09 \\
DietNeRF & SSIM & \cellcolor{Orange!30}0.921 & \cellcolor{Red!30}0.970 & \cellcolor{Red!30}0.949 & \cellcolor{Red!30}0.949 & \cellcolor{Red!30}0.924 & \cellcolor{Orange!30}0.911 & \cellcolor{Yellow!30}0.894 & \cellcolor{Red!30}0.971 & \cellcolor{Red!30}0.936 \\
 & LPIPS & \cellcolor{Yellow!30}0.085 & \cellcolor{Orange!30}0.060 & \cellcolor{Orange!30}0.079 & \cellcolor{Yellow!30}0.103 & \cellcolor{Red!30}0.093 & \cellcolor{Yellow!30}0.166 & \cellcolor{Yellow!30}0.207 & \cellcolor{Orange!30}0.060 & \cellcolor{Yellow!30}0.107 \\
\midrule
 & PSNR & \cellcolor{Orange!30}24.29 & \cellcolor{Orange!30}26.11 & \cellcolor{Orange!30}23.84 & \cellcolor{Yellow!30}22.89 & \cellcolor{Yellow!30}20.06 & \cellcolor{Yellow!30}18.33 & \cellcolor{Yellow!30}13.63 & \cellcolor{Yellow!30}19.60 & \cellcolor{Yellow!30}21.09 \\
InfoNeRF & SSIM & \cellcolor{Red!30}0.923 & \cellcolor{Orange!30}0.961 & \cellcolor{Orange!30}0.944 & \cellcolor{Yellow!30}0.937 & \cellcolor{Yellow!30}0.897 & \cellcolor{Yellow!30}0.877 & \cellcolor{Red!30}0.905 & \cellcolor{Yellow!30}0.962 & \cellcolor{Yellow!30}0.926 \\
 & LPIPS & \cellcolor{Red!30}0.069 & \cellcolor{Red!30}0.059 & \cellcolor{Red!30}0.073 & \cellcolor{Red!30}0.092 & \cellcolor{Yellow!30}0.117 & \cellcolor{Orange!30}0.161 & \cellcolor{Orange!30}0.181 & \cellcolor{Yellow!30}0.094 & \cellcolor{Orange!30}0.106 \\
\midrule
 & PSNR & 12.91 & 11.54 & 10.79 & 11.70 & 10.17 & 11.46 & 11.18 & 17.95 & 12.21 \\
FreeNeRF & SSIM & 0.779 & 0.827 & 0.786 & 0.796 & 0.791 & 0.751 & 0.746 & 0.899 & 0.797 \\
 & LPIPS & 0.210 & 0.312 & 0.351 & 0.461 & 0.220 & 0.554 & 0.458 & 0.299 & 0.358 \\
\midrule
 & PSNR & \cellcolor{Red!30}25.07 & \cellcolor{Yellow!30}26.07 & \cellcolor{Yellow!30}23.72 & \cellcolor{Red!30}26.27 & \cellcolor{Orange!30}20.68 & \cellcolor{Red!30}23.14 & \cellcolor{Red!30}21.91 & \cellcolor{Red!30}29.44 & \cellcolor{Red!30}24.42 \\
Ours & SSIM & \cellcolor{Yellow!30}0.918 & \cellcolor{Orange!30}0.961 & \cellcolor{Yellow!30}0.936 & \cellcolor{Orange!30}0.946 & \cellcolor{Orange!30}0.903 & \cellcolor{Red!30}0.912 & \cellcolor{Red!30}0.905 & \cellcolor{Orange!30}0.965 & \cellcolor{Orange!30}0.930 \\
 & LPIPS & \cellcolor{Orange!30}0.072 & \cellcolor{Yellow!30}0.075 & \cellcolor{Yellow!30}0.089 & \cellcolor{Orange!30}0.096 & \cellcolor{Orange!30}0.116 & \cellcolor{Red!30}0.134 & \cellcolor{Red!30}0.139 & \cellcolor{Red!30}0.058 & \cellcolor{Red!30}0.098 \\
\bottomrule
    \end{tabularx}
\end{table*}

\section{Comparisons on DTU}

We include DTU for the sake of completeness though it is a forward-facing dataset and falls outside our focus of interest.
There are different considerations in sparse view reconstruction for forward-facing and 360 --
for forward-facing scenes and objects, as the back side is undefined, the features are also largely undefined.
In this case, ZeroRF still performs better than or on-par with the state-of-the-art methods (Tab.~\ref{tab:dtu3vfull}), but
does not show a significant margin.

\begin{table*}[tbh]
    \caption{Comparison of per-scene metrics of DTU 3 view settings.}
    \label{tab:dtu3vfull}
    \vspace{-0.75em}
    \centering
    \scriptsize
    \begin{tabularx}{0.98\textwidth}{l c *{15}{Y}}
\toprule
 & Scan & 24 & 37 & 40 & 55 & 63 & 65 & 69 & 83 & 97 & 105 & 106 & 110 & 114 & 118 & 122 \\
\midrule
 & PSNR & \cellcolor{Yellow!30}10.37 & \cellcolor{Yellow!30}13.06 & \cellcolor{Orange!30}12.69 & 12.92 & \cellcolor{Yellow!30}20.24 & \cellcolor{Orange!30}17.99 & \cellcolor{Red!30}17.91 & \cellcolor{Yellow!30}18.85 & \cellcolor{Yellow!30}13.47 & \cellcolor{Yellow!30}14.83 & \cellcolor{Orange!30}19.52 & \cellcolor{Yellow!30}18.04 & \cellcolor{Yellow!30}18.09 & \cellcolor{Orange!30}22.98 & \cellcolor{Yellow!30}23.65 \\
DietNeRF & SSIM & 0.245 & \cellcolor{Yellow!30}0.525 & 0.296 & 0.322 & \cellcolor{Yellow!30}0.810 & \cellcolor{Yellow!30}0.801 & \cellcolor{Yellow!30}0.433 & \cellcolor{Yellow!30}0.702 & \cellcolor{Yellow!30}0.333 & \cellcolor{Yellow!30}0.417 & \cellcolor{Yellow!30}0.693 & \cellcolor{Yellow!30}0.520 & \cellcolor{Yellow!30}0.630 & \cellcolor{Yellow!30}0.771 & \cellcolor{Yellow!30}0.786 \\
 & LPIPS & 0.615 & \cellcolor{Yellow!30}0.372 & \cellcolor{Yellow!30}0.541 & \cellcolor{Yellow!30}0.411 & \cellcolor{Yellow!30}0.219 & \cellcolor{Orange!30}0.198 & \cellcolor{Yellow!30}0.413 & \cellcolor{Yellow!30}0.214 & \cellcolor{Yellow!30}0.432 & \cellcolor{Yellow!30}0.383 & \cellcolor{Yellow!30}0.313 & \cellcolor{Yellow!30}0.320 & \cellcolor{Yellow!30}0.301 & \cellcolor{Red!30}0.228 & \cellcolor{Yellow!30}0.199 \\
\midrule
 & PSNR & 10.32 & 8.34 & 9.25 & \cellcolor{Yellow!30}14.86 & 5.08 & 12.30 & 12.32 & 9.01 & 9.11 & 9.49 & 15.64 & 14.58 & 16.80 & 16.44 & 17.72 \\
InfoNeRF & SSIM & \cellcolor{Yellow!30}0.446 & 0.342 & \cellcolor{Yellow!30}0.407 & \cellcolor{Yellow!30}0.384 & 0.300 & 0.399 & 0.109 & 0.417 & 0.303 & 0.322 & 0.359 & 0.186 & 0.495 & 0.407 & 0.497 \\
 & LPIPS & \cellcolor{Yellow!30}0.564 & 0.505 & 0.556 & 0.567 & 0.605 & 0.549 & 0.561 & 0.571 & 0.569 & 0.568 & 0.481 & 0.495 & 0.451 & 0.460 & 0.451 \\
\midrule
 & PSNR & \cellcolor{Orange!30}10.81 & \cellcolor{Red!30}17.41 & \cellcolor{Yellow!30}11.93 & \cellcolor{Orange!30}16.72 & \cellcolor{Red!30}22.02 & \cellcolor{Red!30}20.37 & \cellcolor{Orange!30}17.02 & \cellcolor{Red!30}28.11 & \cellcolor{Orange!30}18.60 & \cellcolor{Red!30}22.13 & \cellcolor{Red!30}21.65 & \cellcolor{Red!30}20.44 & \cellcolor{Red!30}21.53 & \cellcolor{Red!30}23.56 & \cellcolor{Red!30}26.16 \\
FlipNeRF & SSIM & \cellcolor{Orange!30}0.475 & \cellcolor{Red!30}0.700 & \cellcolor{Orange!30}0.523 & \cellcolor{Orange!30}0.686 & \cellcolor{Orange!30}0.880 & \cellcolor{Red!30}0.865 & \cellcolor{Orange!30}0.640 & \cellcolor{Red!30}0.943 & \cellcolor{Orange!30}0.775 & \cellcolor{Red!30}0.843 & \cellcolor{Red!30}0.800 & \cellcolor{Orange!30}0.819 & \cellcolor{Red!30}0.795 & \cellcolor{Red!30}0.828 & \cellcolor{Orange!30}0.869 \\
 & LPIPS & \cellcolor{Orange!30}0.462 & \cellcolor{Red!30}0.186 & \cellcolor{Orange!30}0.452 & \cellcolor{Orange!30}0.268 & \cellcolor{Orange!30}0.174 & \cellcolor{Red!30}0.160 & \cellcolor{Orange!30}0.332 & \cellcolor{Red!30}0.104 & \cellcolor{Orange!30}0.257 & \cellcolor{Red!30}0.196 & \cellcolor{Red!30}0.294 & \cellcolor{Orange!30}0.209 & \cellcolor{Orange!30}0.267 & \cellcolor{Yellow!30}0.267 & \cellcolor{Orange!30}0.181 \\
\midrule
 & PSNR & \cellcolor{Red!30}14.43 & \cellcolor{Orange!30}15.46 & \cellcolor{Red!30}17.66 & \cellcolor{Red!30}19.06 & \cellcolor{Orange!30}21.19 & \cellcolor{Yellow!30}17.26 & \cellcolor{Yellow!30}16.25 & \cellcolor{Orange!30}23.65 & \cellcolor{Red!30}20.14 & \cellcolor{Orange!30}20.08 & \cellcolor{Yellow!30}18.07 & \cellcolor{Orange!30}20.43 & \cellcolor{Orange!30}19.69 & \cellcolor{Yellow!30}21.14 & \cellcolor{Orange!30}23.72 \\
Ours & SSIM & \cellcolor{Red!30}0.532 & \cellcolor{Orange!30}0.677 & \cellcolor{Red!30}0.590 & \cellcolor{Red!30}0.731 & \cellcolor{Red!30}0.882 & \cellcolor{Orange!30}0.822 & \cellcolor{Red!30}0.730 & \cellcolor{Orange!30}0.922 & \cellcolor{Red!30}0.784 & \cellcolor{Orange!30}0.835 & \cellcolor{Orange!30}0.767 & \cellcolor{Red!30}0.857 & \cellcolor{Orange!30}0.762 & \cellcolor{Orange!30}0.812 & \cellcolor{Red!30}0.884 \\
 & LPIPS & \cellcolor{Red!30}0.370 & \cellcolor{Orange!30}0.188 & \cellcolor{Red!30}0.399 & \cellcolor{Red!30}0.239 & \cellcolor{Red!30}0.135 & \cellcolor{Yellow!30}0.217 & \cellcolor{Red!30}0.309 & \cellcolor{Orange!30}0.118 & \cellcolor{Red!30}0.209 & \cellcolor{Orange!30}0.201 & \cellcolor{Orange!30}0.307 & \cellcolor{Red!30}0.202 & \cellcolor{Red!30}0.251 & \cellcolor{Orange!30}0.246 & \cellcolor{Red!30}0.160 \\

\bottomrule
    \end{tabularx}
\end{table*}

\section{Architecture Implementation}

The SD Decoder generator (final generator for ZeroRF) architecture consists of ResNet convolutional blocks and upsampling modules. More hyperparameters are listed in Tab.~\ref{tab:archgenerator}. The input noise resolutions for NeRF-Synthetic, OpenIllumination and DTU are $20$ while it is $7$ for generation and editing tasks. It is about $1/40$ of the image resolution. The network has only 7M parameters, and the computation is negligible compared to per-point decoding and ray integral.
The decoder architecture is illustrated in Fig.~\ref{fig:archdecoder}, which is a direct implementation of Eq.~(5, 6) in the main paper.

\begin{table*}[tbh]
    \caption{Generator architecture listing.}
    \label{tab:archgenerator}
    \vspace{-0.75em}
    \centering
    \small
    \begin{tabular}{lc}
        \toprule
        Item & Configuration \\
        \midrule
        Input noise channels & 8 \\
        Output feature channels & 16 \\
        Block resolutions & $1\times, 2\times, 4\times, 8\times, 16\times, 16\times$ \\
        ResNet basic blocks per block & $2, 4, 4, 4, 4, 4$ \\
        \# Parameters & $7.0$ M \\
        \bottomrule
    \end{tabular}
\end{table*}

\begin{figure*}[tbh]
    \centering
    \includegraphics[width=0.8\textwidth]{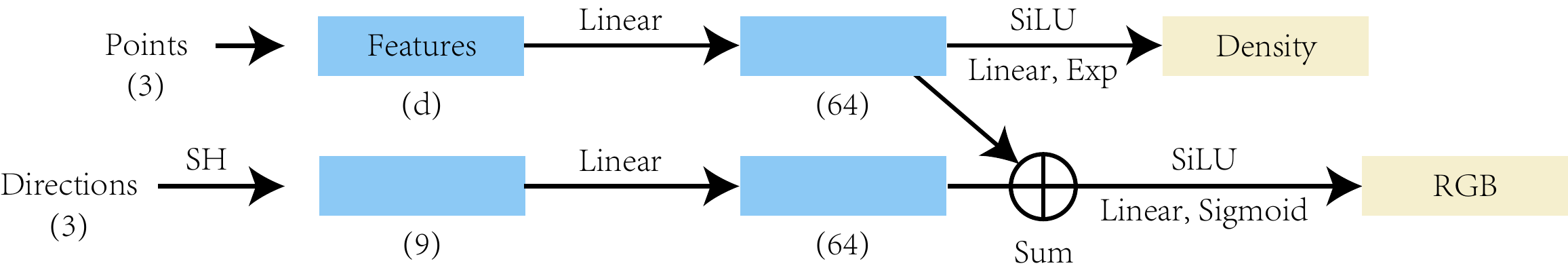}
    \caption{Decoder architecture.}
    \label{fig:archdecoder}
\end{figure*}

\section{Limitations and Future Work}

We discuss more about the limitations and future work of ZeroRF in this section. We found in our experiments that ZeroRF has a chance to magnify the weakness in the underlying representations. For example, it is known that TensoRF exhibits axis-aligned artifacts under SO(3) rotations \cite{gao2023iCCV}. Under certain circumstances, ZeroRF (on TensoRF) will bias towards axis-aligned geometries (see the edges of the hat in Fig.~5 of main paper, as well as the pumpkins in Fig.~\ref{fig:oppo_supp}). Applying ZeroRF to DiF does not have this issue, but minor floaters in unseen areas may occur.

\begin{figure*}
    \centering
    \includegraphics[width=0.5\linewidth]{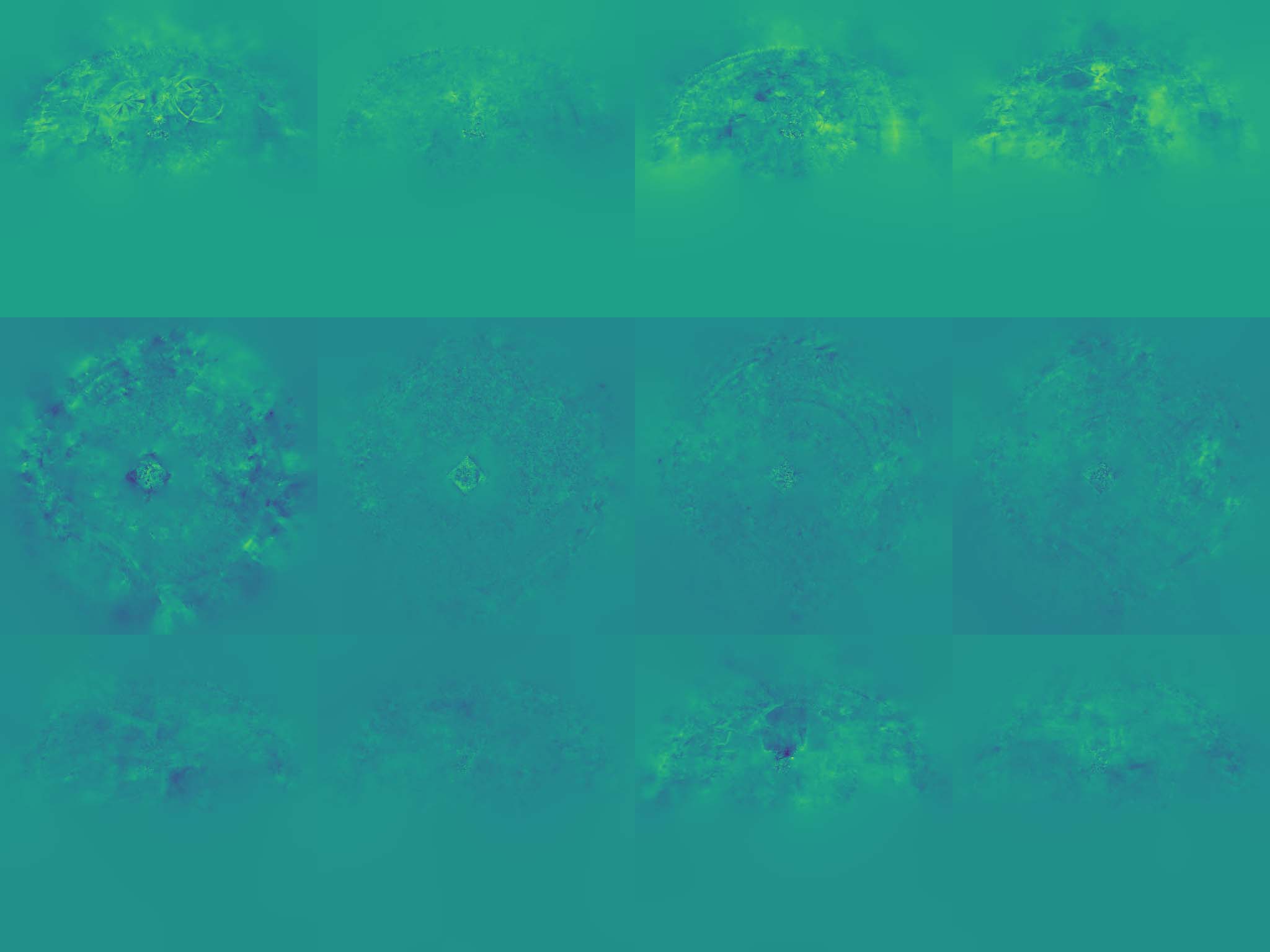}
    \caption{Visualization of features from dense-view TensoRF on the Bonsai scene from the mip-NeRF 360 dataset.}
    \label{fig:bonsai}
\end{figure*}

Another future work for ZeroRF, as mentioned in the main paper, is to apply it for unbounded scenes. Grid representations usually perform a non-linear contraction in space to represent unbounded scenes, which leads to features being distorted, especially for the background areas. The features are thus hardly perceivable as a natural image, as shown in Fig.~\ref{fig:bonsai}. Consequently, extra work would be needed to apply our technique to unbounded scenes.

\begin{figure*}[th]
\begin{center}
   \includegraphics[width=\linewidth]{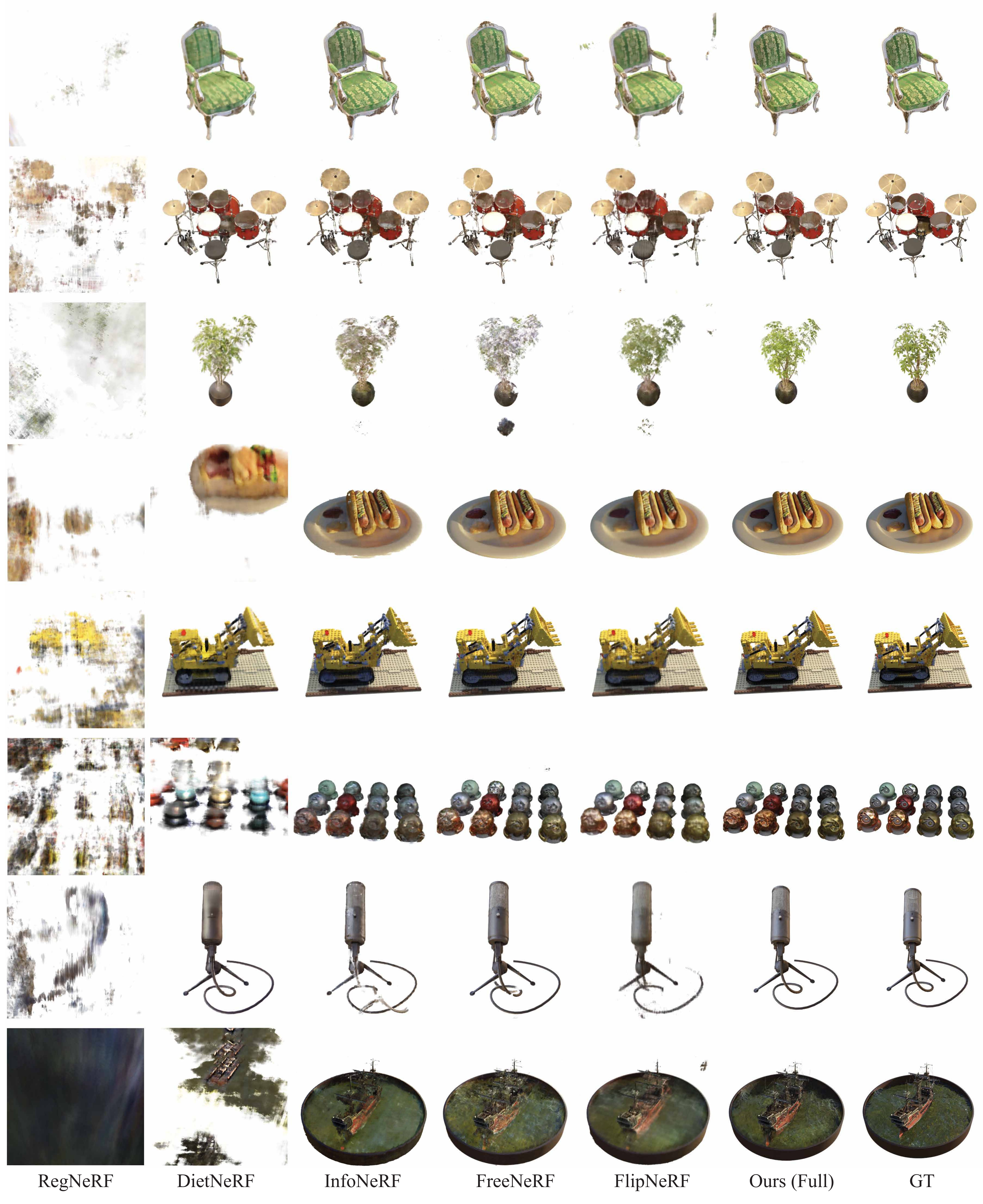}
\end{center}
\vspace{-1.5em}
  \caption{Per-scene qualitative comparisons of NeRF-Synthetic 6 view settings.}
\label{fig:nerfsyn_supp}
\end{figure*}

\begin{figure*}[th]
\begin{center}
   \includegraphics[width=0.6\linewidth]{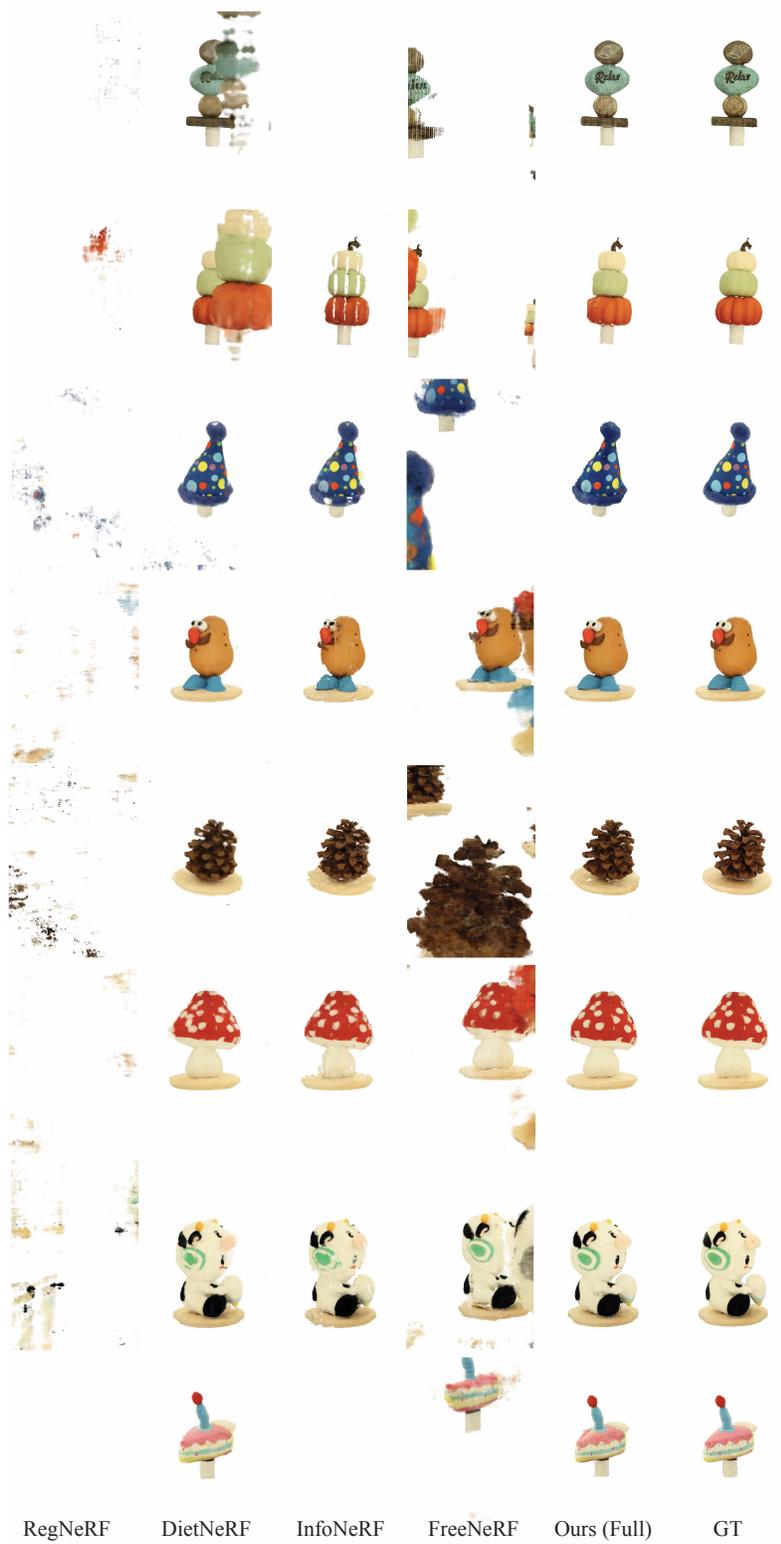}
\end{center}
\vspace{-1.5em}
  \caption{Per-scene qualitative comparisons of OpenIllumination 6 view settings.}
\label{fig:oppo_supp}
\end{figure*}
\fi


\end{document}